\documentclass[acmtog,nonacm,screen,authorversion]{acmart}

\AtBeginDocument{%
  }

\setcopyright{acmlicensed}
\copyrightyear{2018}
\acmYear{2018}
\acmDOI{XXXXXXX.XXXXXXX}
\acmISBN{978-1-4503-XXXX-X/2018/06}
\usepackage{multirow}
\usepackage{pifont}
\usepackage[most]{tcolorbox}
\usepackage{xcolor}
\definecolor{darkgreen}{rgb}{0.0, 0.8, 0.0}

\tcbuselibrary{listings,breakable,skins}
\newtcblisting{promptbox}[2][]{
  enhanced,
  breakable,
  width=\linewidth,
  colback=white,
  colframe=black!70,
  coltitle=white,
  colbacktitle=black!80,
  title={#2},
  fonttitle=\bfseries\footnotesize,
  boxrule=0.6pt,
  arc=1.5mm,
  left=1.5mm,
  right=1.5mm,
  top=0.8mm,
  bottom=0.8mm,
  before skip=4pt,
  after skip=6pt,
  listing only,
  listing options={
    basicstyle=\ttfamily\scriptsize,
    breaklines=true,
    columns=fullflexible,
    keepspaces=true,
    showstringspaces=false
  },
  #1
}
\newcommand{\cmark}{\textcolor{darkgreen}{\ding{52}}}
\newcommand{\xmark}{\textcolor{red}{\ding{55}}}
\newcommand{\pmark}{$\triangle$}

\acmSubmissionID{1804}

\begin{document}


\title{Feedforward 3D Editing Learns from Semantic-Part Transformation}

\author{Jiawei Weng}
\authornote{Equal contribution.}
\email{jweng007@e.ntu.edu.sg}
\affiliation{%
  \institution{Nanyang Technological University}
  \country{Singapore}
}

\author{Saining Zhang}
\authornotemark[1]
\authornote{Corresponding author.}
\email{saining002@e.ntu.edu.sg}
\affiliation{%
  \institution{Nanyang Technological University}
  \country{Singapore}
}

\author{Zhenxin Diao}
\authornotemark[1]
\email{diaozhenxin2005@outlook.com}
\affiliation{%
  \institution{Tsinghua University}
  \country{China}
}

\author{Peishuo Li}
\email{peishuo001@e.ntu.edu.sg}
\affiliation{%
  \institution{Nanyang Technological University}
  \country{Singapore}
}

\author{Henghaofan Zhang}
\email{hhfzhang@outlook.com}
\affiliation{%
  \institution{Tsinghua University}
  \country{China}
}

\author{Junhao Chen}
\email{yisuanwang@gmail.com}
\affiliation{%
  \institution{Tsinghua University}
  \country{China}
}

\author{Hao Zhao}
\authornotemark[2]
\email{zhaohao@air.tsinghua.edu.cn}
\affiliation{%
  \institution{Tsinghua University}
  \country{China}
}

\renewcommand{\shortauthors}{Weng et al.}

\begin{abstract}
3D editing is a fundamental capability for scalable 3D content creation. While image editing has rapidly evolved toward large-scale feedforward generative paradigms, 3D AI generation remains dominated by training-free editing pipelines. A central challenge of feedforward 3D editing lies in the lack of high-quality paired supervision. Editable 3D assets require simultaneous preservation of geometry, multi-view consistency, structural coherence, and localized edit controllability. Existing 3D editing datasets often rely on independently generated assets, image-mediated reconstruction or narrow edit taxonomies, leading to inaccurate localization, weak preservation, blurred edit boundaries, and limited semantic consistency. In this work, we introduce a new perspective: scalable feedforward 3D editing should be learned from semantic-part transformations. Based on this insight, we propose Pxform, a high-quality 3D editing dataset with over 100K consistent before/after editing pairs across seven edit types. Instead of treating objects as unstructured shapes, our pipeline grounds edits directly in semantic 3D parts. Built upon Pxform, we further propose PartFlow, a feedforward 3D editing network that injects source-aware latent control into pretrained 3D generative priors. PartFlow introduces mask-aware velocity preservation and render-space consistency supervision to jointly improve edit fidelity and source preservation, while requiring no 3D edit mask during inference. Extensive experiments demonstrate that high-quality semantic-part supervision substantially improves scalable 3D editing, enabling PartFlow to achieve state-of-the-art performance on both geometric and appearance editing benchmarks. Project page: \href{https://dennis-jwweng.github.io/pxform/}{\color{magenta} {https://dennis-jwweng.github.io/pxform/}}
\end{abstract}

\ccsdesc[500]{Computing methodologies~Computer vision}

\keywords{image-to-3D models, 3D editing, diffusion models}
\begin{teaserfigure}
 \centering
  \includegraphics[width=1.0\textwidth]{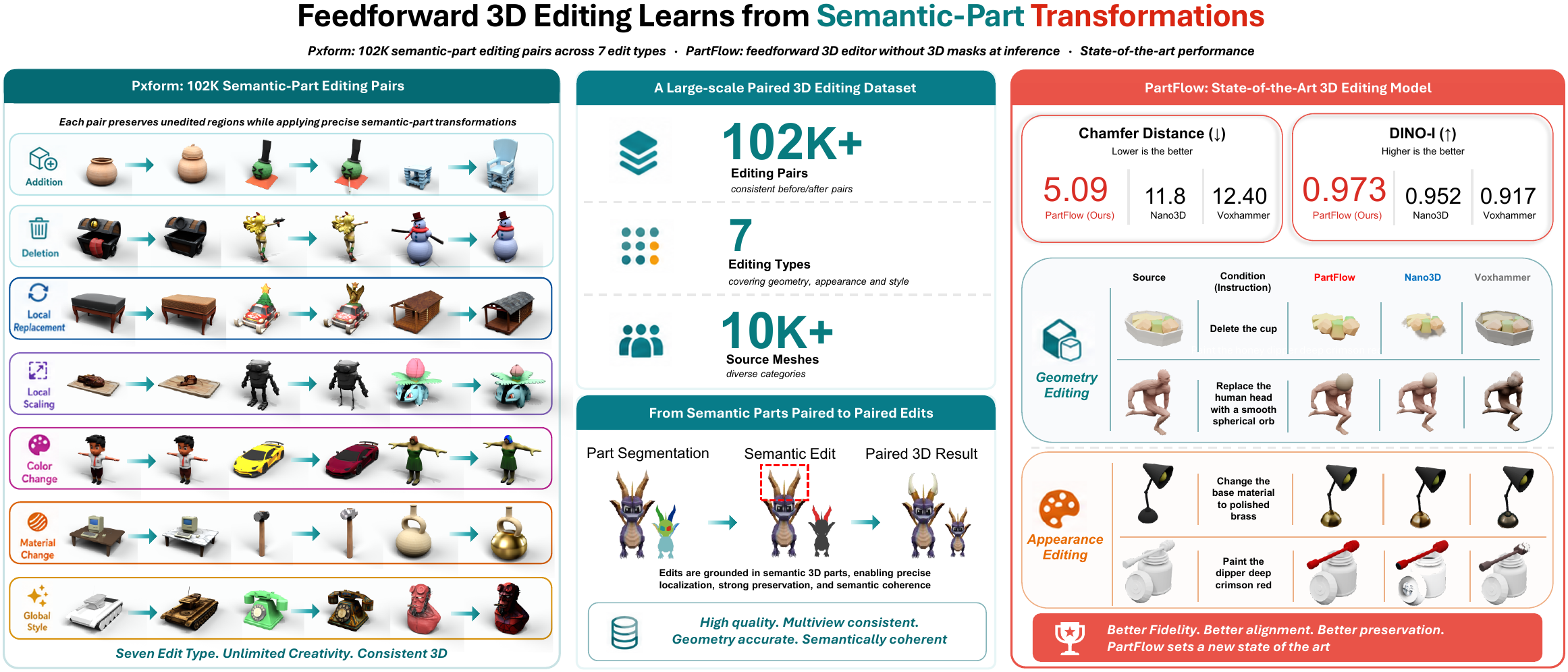}
\caption{We introduce Pxform, a high-quality holistic 3D editing dataset with over 100K consistent before/after pairs, covering seven edit types: addition, deletion, local replacement, local scaling, local color change, local material change, and global style transformation. By grounding edits in semantic 3D parts, Pxform provides the paired supervision needed for learning feedforward 3D editing from semantic-part transformations, and enables PartFlow to scale from isolated examples toward general-purpose 3D editing.
}
  \label{fig:teaser}
\end{teaserfigure}


\maketitle
\begin{table*}[t]
  \caption{Comparison of 3D editing datasets. "Region" denotes whether localized edit regions are available or recoverable.
"Add/Del." denotes local addition and deletion.
"Modif." includes local replacement and scaling edits.
"T/M" denotes local texture or material editing.
"Style" denotes style-level appearance transformation.
"Consist." indicates whether unedited regions are strictly preserved.
"Harmony" indicates whether the edited regions remain semantically coherent with the preserved regions.
$\triangle$ denotes only in test set.
Pose-change samples are excluded from the reported statistics.}
  \label{tab:dataset_comparison}
  \centering
  \begin{tabular*}{\textwidth}{@{\extracolsep{\fill}}l|cc|c|cccc|cc}
    \toprule
    \multirow{2}{*}{Dataset} &
    \multicolumn{2}{c|}{Scale} &
    \multirow{2}{*}{Region} &
    \multicolumn{4}{c|}{Edit Coverage} &
    \multicolumn{2}{c}{Pair Quality} \\
    \cline{2-3}
    \cline{5-8}
    \cline{9-10}
    &
    Train &
    Test &
    &
    Add/Del. &
    Modif. &
    T/M &
    Style &
    Consist. &
    Harmony \\
    \midrule

    3D-Alpaca-Editing~\cite{ye2025shapellmomni}
    & 52,532 & -- 
    & \xmark & \cmark & \cmark & \xmark & \xmark & \xmark & \cmark \\

    CMD~\cite{li2025cmd}
    & 40,000 & 50
    & \xmark & \cmark & \xmark & \xmark & \xmark & \cmark & \xmark \\

    Edit3D-Bench~\cite{li2025voxhammer}
    & -- & 300
    & \cmark & \xmark & \cmark & \xmark & \xmark & \cmark & \cmark \\

    3DEditVerse~\cite{xia2025scalable3dediting}
    & 80,403 & 814
    & \cmark & \cmark & \cmark & \xmark & \xmark & \cmark & \cmark \\

    Steer3D-Data~\cite{ma2025steer3d}
    & 320K & 250
    & \xmark & \cmark & \cmark & \cmark & \xmark & \xmark & \cmark \\

    ShapeUP-Parts~\cite{gat2026shapeup}
    & 13.7K--27.5K & 100
    & \cmark & \cmark & \xmark & \pmark & \pmark & \cmark & \cmark \\

    Edit-3DVerse~\cite{xu2026beyondvoxel}
    & 100K & 100
    & \xmark & \cmark & \cmark & \cmark & \cmark & \xmark & \cmark \\

    Nano3D-Edit-100K~\cite{ye2025nano3d}
    & 100K & -
    & \xmark & \cmark & \cmark & \xmark & \xmark & \cmark & \cmark \\

    \midrule

    Pxform (Ours)
    & 102,007 & 1,497
    & \cmark & \cmark & \cmark & \cmark & \cmark & \cmark & \cmark \\

    \bottomrule
  \end{tabular*}
  \footnotesize
\end{table*}

\section{Introduction}

3D editing modifies the geometry or appearance of an existing 3D asset according to user input while preserving its identity, structure, and unedited regions. It is essential for scalable 3D content creation, game and film production, digital twins, and AR/VR authoring, where assets are refined through iterative design workflows. Recent 2D image editing models~\cite{blackforestlabs2025fluxkontext,google2025gemini25flashimage,qwen2025imageedit,chow2026weave,wang2025selftok,jia2026imagine} show that instruction-driven data can scale generative editing priors. However, extending this paradigm to 3D is harder: a 3D editor must jointly satisfy instruction following, 3D edit localization, cross-view consistency, geometric fidelity, and source preservation across diverse categories.

Similar to the early evolution of image and video editing, recent 3D editing first advanced through training-free paradigms such as inversion-trajectory editing~\cite{mokady2023nulltext} and FlowEdit~\cite{kulikov2024flowedit}. While effective in selected cases, these methods remain hard to scale: some require manually specified 3D edit regions~\cite{li2025voxhammer}, some rely on unstable latent merging~\cite{ye2025nano3d}, and others use long agentic pipelines where errors may accumulate~\cite{chi2026vinedresser3d}. These limitations make training-free methods less suitable as a foundation for general-purpose 3D editing.

\begin{figure*}[t]
  \centering
  \includegraphics[width=1.0\textwidth]{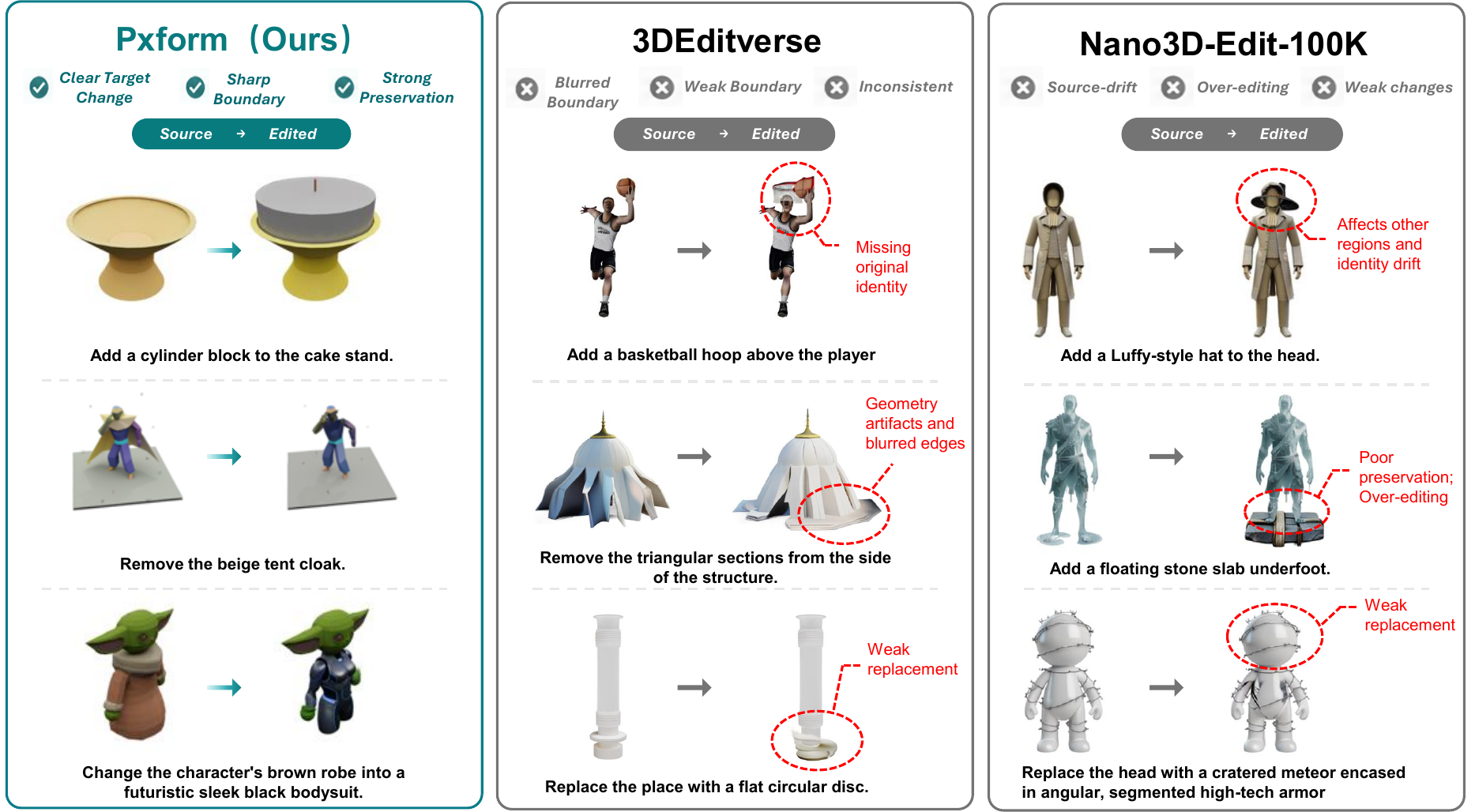}
  \caption{
    Qualitative comparison of editing pairs from Pxform, 3DEditVerse \cite{xia2025scalable3dediting}, and Nano3D-Edit-100K \cite{ye2025nano3d}. Pxform provides part-semantic edits with clear target changes, sharp edit boundaries, and strong preservation of unedited regions. In contrast, other datasets may exhibit weak edit execution, blurred local geometry, or inconsistent preservation.
    }
    \Description{
    A qualitative comparison of before-and-after 3D editing examples from H3D, 3DEditVerse, and Nano3D-Edit-100K.
    }
  \label{fig:data_comparison}
\end{figure*}

Recent training-based methods learn 3D edit transformations from paired data, synthetic triplets, or instruction-tuned supervision~\cite{ma2025steer3d,gat2026shapeup,xia2025scalable3dediting,xu2026beyondvoxel,ye2025shapellmomni,ye2026omni123,huang2026unimesh,yang2026eva01}. However, as shown in Table~\ref{tab:dataset_comparison}, their scalability is primarily constrained by data quality, diversity, and consistency. Existing datasets either rely on independently generated or image-mediated before/after assets~\cite{ye2025shapellmomni,ma2025steer3d,xu2026beyondvoxel}, focus mainly on simple part addition or deletion~\cite{gat2026shapeup,li2025cmd}, or are synthesized by training-free pipelines with limited region control. For example, Nano3D-Edit-100K~\cite{ye2025nano3d} lacks explicit region constraints, while 3DEditVerse~\cite{xia2025scalable3dediting} depends on multi-view 2D segmentation, which may lead to inaccurate localization, blurred edit boundaries, and weaker preservation, as shown in Figure~\ref{fig:data_comparison}.

To address this data bottleneck, we introduce \textbf{Pxform}, a large-scale and comprehensive part-semantic 3D editing dataset designed to supervise feedforward 3D editing through consistent semantic-part transformations.
Pxform contains 102,007 training pairs and 1,497 test pairs from 11,273 curated source meshes, covering seven edit types: addition, deletion, local replacement, local scaling, local color change, local material change, and global style transformation.
To construct Pxform, we develop an agent-assisted part-semantic data engine that grounds each edit in the source mesh's semantic parts, rather than treating the asset as an unstructured object or relying only on 2D masks.
By refining part semantics, planning edit targets, and verifying results with multi-view visual reasoning, our pipeline produces edits that are semantically aligned, spatially localized, and structurally consistent. Combined with explicit 3D part-mask control, this construction yields cleaner boundaries, stronger preservation of unedited regions, and broader coverage, as shown in Table~\ref{tab:dataset_comparison} and Figure~\ref{fig:data_comparison}.

Built on Pxform, we further propose \textbf{PartFlow}, a ControlNet-style 3D editing model for structure-preserving edit generation. PartFlow injects source-latent information through a lightweight control branch, uses training-only edit masks to suppress off-target changes, and adds render-space supervision to improve visual alignment and condition controllability. Trained on Pxform, PartFlow achieves state-of-the-art performance on \textbf{Uni3DEdit-Bench}, the 1,497-sample Pxform test split covering both shape and appearance editing tasks, demonstrating the effectiveness of high-quality holistic data for scalable 3D editing.
Our contributions are summarized as follows:
\begin{itemize}
    \item We build \textbf{Pxform}, a large-scale semantic-part transformation dataset for 3D editing, with 102,007 training pairs and 1,497 test pairs from 11,273 curated source meshes across seven edit types.
    \item We propose an agent-assisted part-semantic data construction pipeline that enables semantic target selection, explicit 3D region control, and multi-view quality verification.
    \item We introduce \textbf{PartFlow}, a feedforward 3D editing model that learns from Pxform’s semantic-part transformations through source-latent control, achieving state-of-the-art performance on \textbf{Uni3DEdit-Bench} while preserving unedited regions without requiring 3D masks at inference.
\end{itemize}

\section{Related Work}

\subsection{3D Generative Models}

3D generative models~\cite{zhang2023shape2vecset,xiang2025structured,hunyuan3d2025hunyuan3d,ye2025hi3dgen,wu2025direct3ds2,li2025sparc3d,chen2025ultra3d,lai2025lattice,jia2025ultrashape,xiang2025native,shen2025gamba,yi2024mvgamba,wu2024consistent3d,yi2024diffusion,xu2024instantmesh,tang2024lgm,chen20253dtopia,li2026pixal3d,li2024craftsman3d,li2025near,xiang2026dvd,poole2022dreamfusion,lin2023magic3d,wang2023sjc,wang2023prolificdreamer,chen2023fantasia3d,metzer2023latentnerf,michel2022text2mesh,sanghi2022clipforge,gao2022get3d,nichol2022pointE,jun2023shapee,tang2023dreamgaussian,hong2023lrm,he2023openlrm,tochilkin2024triposr,liu2023zero123,shi2023zero123pp,shi2023mvdream,liu2023syncdreamer,long2024wonder3d,zhang2024clay,hyper3drodin2026} have evolved from point-, mesh-, and neural-field-based representations toward scalable latent diffusion and flow-based generation. Early works such as 3DShape2VecSet~\cite{zhang2023shape2vecset} encode shapes as vector sets for transformer-based diffusion modeling, while recent systems build stronger structured or sparse 3D latents with image/text conditioning, including TRELLIS~\cite{xiang2025structured}, Hunyuan3D 2.1~\cite{hunyuan3d2025hunyuan3d}, and Hi3DGen~\cite{ye2025hi3dgen}. Further advances improve resolution, efficiency, and geometry through sparse volumes, compact native latents, modality-consistent VAEs, and part-aware refinement~\cite{wu2025direct3ds2,li2025sparc3d,chen2025ultra3d,lai2025lattice,jia2025ultrashape,xiang2025native}. Despite providing strong 3D priors, these models mainly target one-shot generation or reconstruction, whereas 3D editing requires modifying an existing asset while preserving its identity and unedited regions.

\subsection{3D Editing}
Early 3D editing methods transfer 2D editing priors to 3D through view-space editing, multi-view propagation, or optimization-based reconstruction~\cite{sella2023voxe,barda2025instant3dit,baron2025editp23,wang2024view,haque2023instructnerf2nerf,zhuang2023dreameditor,wang2023nerfart,chen2024gaussianeditor,fang2024gaussiangrouping}. While effective for some edits, they depend on projection and reconstruction consistency, making precise 3D-localized editing difficult.
With structured 3D generative models, recent training-free methods edit in 3D latent spaces by reusing frozen priors through inversion, latent replacement, flow-based editing, or agentic tool chains~\cite{li2025voxhammer,ye2025nano3d,zhou2025anchorflow,hu2026easy3e,chi2026vinedresser3d,chen2025idea23d,liu2026velocity}. These methods reveal the editability of pretrained 3D priors, but remain method-specific and unstable across diverse objects, edit types, and instructions.
Training-based methods learn 3D edit transformations from paired data, synthetic triplets, or instruction-tuned supervision. Tailor3D \cite{qi2024tailor3d} and BVE \cite{xu2026beyondvoxel} perform image-conditioned 3D customization or editing, but may suffer from source drift without strong original-latent reference control. Dataset-driven methods such as 3DEditFormer \cite{xia2025scalable3dediting} and ShapeUP \cite{gat2026shapeup} adapt pretrained 3D generative backbones using paired editing data, while ShapeLLM-Omni \cite{ye2025shapellmomni}, Omni123 \cite{ye2026omni123}, and UniMesh \cite{huang2026unimesh} explore unified models for 3D understanding, generation, and editing. These works show the promise of training-based 3D editing, but remain limited by the quality, diversity, edit coverage, and structural consistency of available data.

\begin{figure*}[t]
  \centering
  \includegraphics[width=1.0\textwidth]{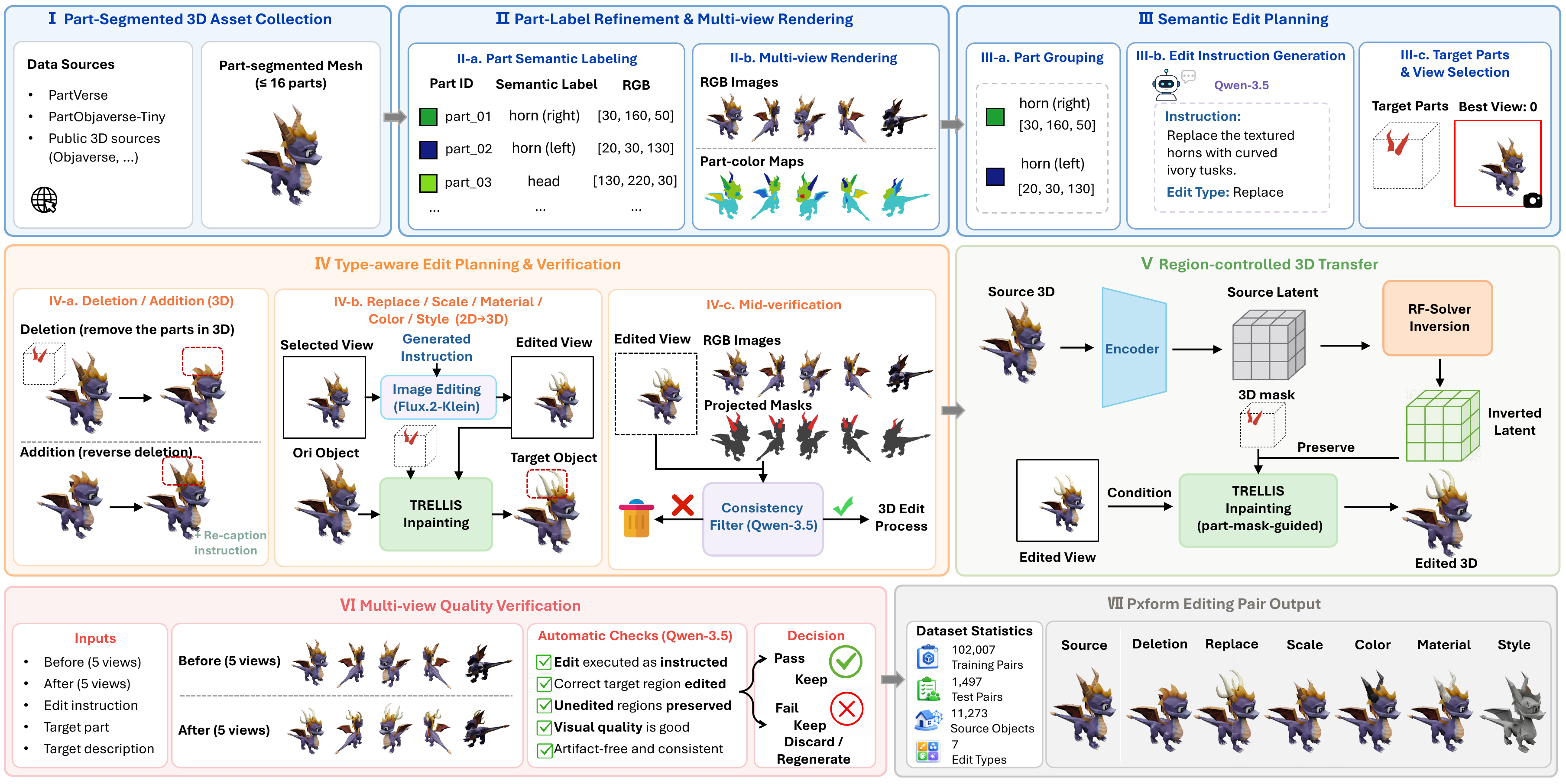}
  \caption{
Overview of the modular Pxform data construction pipeline. Starting from part-segmented 3D assets, the pipeline first refines semantic part labels and renders multi-view RGB images and part-color maps for semantic grounding and cross-view correspondence. An agent-assisted semantic planner then groups semantic parts, generates edit instructions, and selects target parts and informative editing views. Type-aware edit planning handles deletion and addition directly in 3D, while replacement, scaling, color, material, and style edits are first converted into edited 2D visual references and subsequently verified through multi-view consistency filtering. The edited conditions are then transferred back into 3D via RF-Solver-based latent inversion and part-mask-guided TRELLIS inpainting, preserving regions outside the target mask. Finally, multi-view quality verification evaluates edit execution, target-region correctness, source preservation, and visual consistency before accepted samples are added into the final Pxform editing-pair dataset.
}
  \Description{
  A pipeline diagram showing how H3D constructs paired 3D editing data from part-segmented meshes through semantic planning, edit construction, 3D part-mask control, and multi-view quality filtering.
  }
  \label{fig:h3d_pipeline}
\end{figure*}

\subsection{3D Edit Datasets}


Large-scale paired 3D editing data remains a major bottleneck for training robust 3D editing models. Existing datasets make early attempts but still have clear limitations. 3D-Alpaca from ShapeLLM-Omni~\cite{ye2025shapellmomni} includes 3D editing instructions, but its before/after pairs are mainly generated through image-mediated processes and may not strictly preserve unchanged regions. CMD~\cite{li2025cmd} and ShapeUP \cite{gat2026shapeup} exploit part-level data to construct editing pairs, but their supervision is largely centered on component addition or deletion, leaving fine-grained local modification underrepresented. More recent datasets such as 3DEditVerse~\cite{xia2025scalable3dediting}, Nano3D-Edit-100K~\cite{ye2025nano3d}, and Edit-3DVerse~\cite{xu2026beyondvoxel} further scale paired 3D editing data, yet they still rely on image-mediated reconstruction, training-free generation, coarse localization, or post-hoc preservation strategies. In contrast, Pxform treats 3D editing data as semantic-part transformations, producing consistent, controllable, and structurally grounded edit pairs with broader geometry- and appearance-level coverage.

\section{Pxform}

\subsection{Design Motivation}



Training scalable 3D editing models requires paired data that is large, diverse, and consistent in edit locality and source preservation. Existing datasets rarely satisfy these requirements simultaneously: some rely on independently generated or image-mediated before/after assets, making unedited-region preservation difficult; some part-based datasets mainly cover simple addition or deletion; and others depend on 2D masks or multi-view projections, leading to inaccurate localization and blurred edit boundaries. These limitations motivate a data construction process grounded directly in 3D part semantics \cite{mo2019partnet,xiang2020sapien,zhang2025bang,wang2025partnext,team2026hy3dbench,liu2023partslip,zhou2023partslippp,zhou2024pointsam,yang2024sampart3d,ma2024findanypart,liu2025partfield,ma2025p3sam,zhu2025partsam,chen2025partgen,liu2024part123,yan2025xpart,yang2025omnipart}. 

To this end, we introduce \textbf{Pxform}, a dataset of semantic-part transformations for high-quality geometry, appearance, and operation-level 3D editing. Pxform provides high-quality, consistent, and semantically grounded supervision for training-based 3D editing, containing 102,007 training pairs and 1,497 test pairs from 11,273 curated source meshes across seven edit types: addition, deletion, local replacement, local scaling, local color change, local material change, and global style transformation.

\subsection{Agent-assisted Part-semantic Data Engine}
Figure~\ref{fig:h3d_pipeline} illustrates the modular Pxform data construction pipeline. The central idea is to construct 3D editing pairs from explicit semantic-part transformations. Instead of treating a 3D asset as an unstructured shape or relying only on post-hoc 2D masks, our pipeline grounds each edit in the source mesh's semantic 3D parts, allowing the target region, edit instruction, editing view, and 3D inpainting mask to be jointly specified.

\paragraph{Part-segmented asset collection.}
We first collect part-segmented 3D assets from PartVerse~\cite{dong2025partverse}, PartObjaverse-Tiny~\cite{yang2024sampart3d}, and other public 3D sources~\cite{deitke2023objaverse}. To ensure reliable semantic grounding and controllable editing, we retain objects with no more than 16 parts. This filtering avoids overly fragmented meshes whose part labels are ambiguous or difficult to associate with a localized edit instruction. Each retained object provides a source mesh together with part-level segmentation, which serves as the structural basis for all subsequent edit construction.

\paragraph{Part-label refinement and multi-view rendering.}
For each source asset, we refine or caption its semantic part labels and assign each part a unique color identifier. We then render multi-view RGB images and corresponding part-color maps from fixed camera views. The RGB renderings provide visual context for instruction generation, while the part-color maps establish cross-view correspondence between semantic labels, color-coded regions, and 3D part masks. This step produces a structured part description containing the part index, semantic label, and RGB code, which is later used to select target regions and project masks across views.

\paragraph{Semantic edit planning.}
Given the refined part labels and multi-view renderings, an agent-assisted semantic planner performs three operations: part grouping, edit instruction generation, and target-view selection. First, semantically related components are grouped into a coherent editable unit, such as pairing the left and right horns. Second, the planner generates a natural-language edit instruction conditioned on the object category, target part semantics, and feasible edit type. Third, it selects the target part colors and the most informative view for 2D visual editing. The output of this stage is a structured editing condition containing the edit type, edit prompt, target semantic parts, part RGB codes, and selected view.

\paragraph{Type-aware edit planning and intermediate verification.}
The planned condition is then dispatched to type-specific edit construction branches. For deletion, we directly remove the selected semantic parts from the 3D structure, producing a localized geometry edit with exact part-level control. Addition samples are constructed by reversing valid deletion pairs and re-captioning them with Qwen-3.5~\cite{qwen2026qwen35} to obtain natural addition instructions. For replacement, scaling, color, material, and style edits, the selected view and edit instruction are sent to FLUX.2 [klein]~\cite{blackforestlabs2026flux2klein} to synthesize an edited 2D visual reference. Before transferring the result back to 3D, we perform intermediate consistency filtering using the edited view, multi-view RGB renderings, and projected masks. This step removes failed 2D edits whose target region is incorrect, whose edit is not visually executed, or whose appearance is inconsistent with the planned instruction.

\paragraph{Region-controlled 3D transfer.}
For valid 2D edit references, we transfer the edit back into 3D through a region-controlled reconstruction process. Specifically, the source asset is encoded into the TRELLIS latent space, and RF-Solver~\cite{wang2024rfsolver} is used for rectified-flow \cite{liu2022rectifiedflow} inversion. The selected 3D part mask is then used to constrain TRELLIS inpainting so that only the target region is modified while regions outside the mask are preserved. The edited 2D view serves as the visual condition, and the semantic part mask provides spatial control. This avoids the drift caused by reconstructing an entirely new 3D asset from a single edited image, and better preserves the source identity, geometry, and unedited appearance.

\paragraph{Multi-view quality verification.}
Finally, we apply a multi-view quality gate to all generated before/after pairs. Given the edit instruction, target-part metadata, source renderings, edited renderings, and target description, the edit filter checks whether the requested edit is correctly executed, whether the modified region matches the selected semantic part, whether unedited regions are preserved, and whether the final asset is visually plausible and artifact-free. Samples that pass the verification are kept, while failed cases are discarded or regenerated. This final filtering stage improves edit locality, before/after consistency, and visual quality.

\paragraph{Dataset output.}
The accepted samples are collected into Pxform as consistent semantic-part editing pairs. The resulting dataset contains geometry-level edits, including addition, deletion, replacement, and scaling, as well as appearance-level edits, including color, material, and style transformations. This construction process enables Pxform to provide cleaner edit boundaries, stronger source preservation, and broader edit-type coverage than image-mediated pseudo-edit datasets. Figure~\ref{fig:teaser} and Figure~\ref{fig:h3d_cases} show representative cases and statistics of Pxform.

\section{PartFlow}

Although Pxform provides consistent paired supervision, 3D editing still requires balancing edit fidelity with source preservation. Directly fine-tuning a pretrained 3D backbone can entangle these objectives, causing geometry drift or unintended changes in unedited regions. Inspired by ControlNet-style conditional generation~\cite{zhang2023adding} and recent controllable 3D generation/editing models~\cite{ma2025steer3d,cao2025physxanything}, we propose \textbf{PartFlow}, a source-latent controlled 3D editing model built on the two-stage TRELLIS representation.

\subsection{Architecture}



\begin{figure*}[t]
  \centering
  \includegraphics[width=0.95\textwidth]{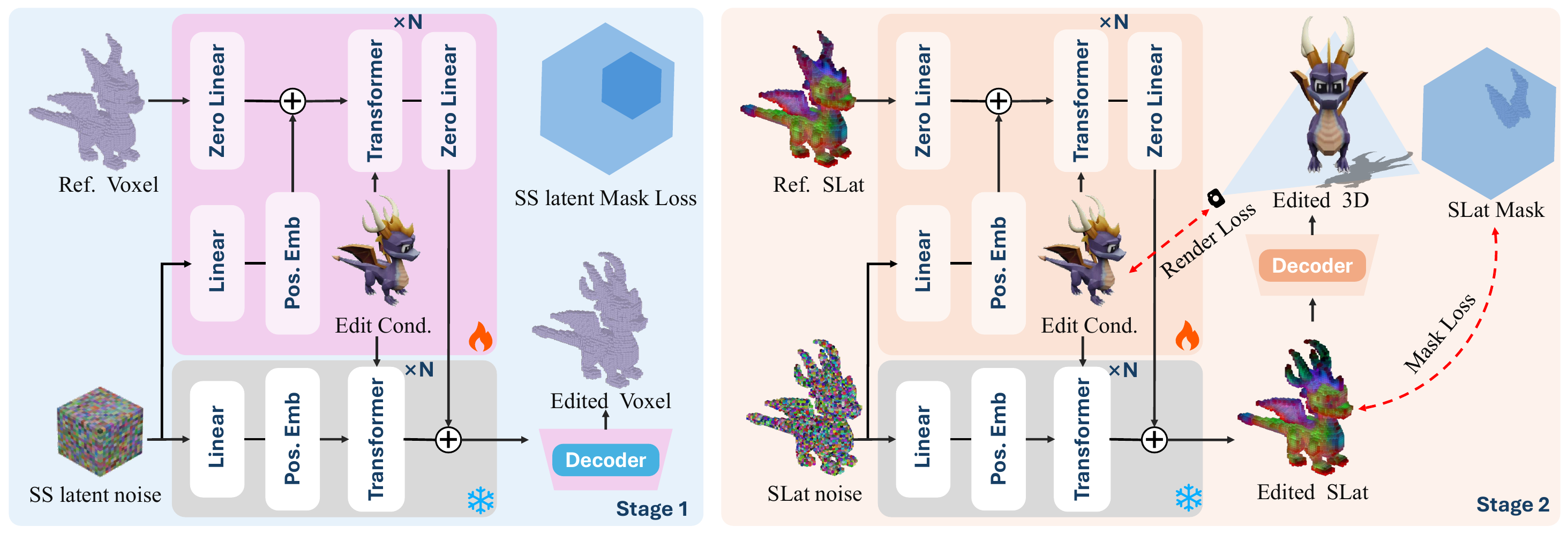}
  \caption{
Overview of PartFlow. PartFlow introduces ControlNet-style source-latent injection into the two-stage TRELLIS editing process: Stage 1 controls coarse sparse-structure editing, while Stage 2 refines SLat-level geometry and appearance. During training, ground-truth edit masks impose a velocity-space preservation loss on unedited regions, while edited regions are supervised by the standard flow objective. A Stage-2 render-space loss further aligns the Gaussian-rendered output with the target editing view.
}
  \Description{
  A two-stage ControlNet-style 3D editing architecture. The first stage edits sparse-structure latents with source voxel control, and the second stage edits SLat representations with source SLat control, mask-aware losses, and render-space supervision.
  }
  \label{fig:PartFlow}
\end{figure*}

As shown in Figure~\ref{fig:PartFlow}, PartFlow follows the two-stage TRELLIS editing process: the sparse-structure stage edits coarse voxel-level geometry, and the SLat stage refines fine-grained geometry and appearance. In both stages, we attach a trainable ControlNet-style branch that takes the source latent as condition and injects source-aware residuals into the pretrained flow backbone through zero-initialized projections. This enables PartFlow to reuse the pretrained 3D prior while preserving source-asset information during editing.

Given a noisy latent $z_t$, timestep $t$, edit condition $c$, and source latent $z^s$, the edited flow field is predicted as
\begin{equation}
v_{\theta}(z_t, t, c, z^s)
=
F_{\theta}(z_t, t, c)
+
C_{\phi}(z^s, t, c),
\end{equation}
where $F_{\theta}$ is the pretrained flow backbone and $C_{\phi}$ is the trainable source-control branch. This formulation is applied to both sparse-structure and SLat stages.




\subsection{Mask-aware Velocity Preservation}
PartFlow does not require a 3D mask at inference time. However, since Pxform is constructed from semantic 3D parts, each training pair naturally provides a ground-truth edit mask. This allows us to use the mask as training-only supervision to preserve unedited regions.

After the initial 40K-step training stage, we introduce a mask-aware velocity preservation loss. Different from applying reconstruction loss directly on latents, we impose the preservation constraint in the flow-velocity space. Let $v_{\theta}$ denote the predicted velocity, $v^{e}$ the standard velocity target toward the edited latent, and $v^{s}$ the velocity target toward the source latent under the same noisy sample. The standard flow-matching \cite{lipman2023flowmatching} loss is:
\begin{equation}
\mathcal{L}_{\mathrm{flow}}
=
\left\| v_{\theta} - v^{e} \right\|_2^2 .
\end{equation}

Let $M_{\mathrm{pres}} \in \{0,1\}$ denote the preservation mask, where $M_{\mathrm{pres}}=1$ corresponds to regions outside the edited part and $M_{\mathrm{pres}}=0$ corresponds to the edited region. We add an additional preservation loss:
\begin{equation}
\mathcal{L}_{\mathrm{mask}}
=
\left\|
M_{\mathrm{pres}} \odot
\left(v_{\theta} - v^{s}\right)
\right\|_2^2 .
\end{equation}
This term encourages the predicted velocity in unedited regions to follow the source asset rather than the edited target, thereby suppressing unintended geometry or appearance drift. The edited region is still supervised by the standard flow loss, so the model remains free to learn the intended modification. We apply this supervision to both the sparse-structure stage and the SLat stage using masks projected to the corresponding latent resolutions.

\subsection{Render-space Consistency Loss}

The mask-aware velocity preservation loss constrains the 3D latent trajectory, but it does not directly guarantee that the edited asset matches the visual editing condition from the selected view. To improve condition controllability, we introduce a render-space consistency loss in Stage 2. Since the SLat representation can be decoded into a 3D Gaussian \cite{kerbl2023gaussian,yu2024mip,wang2026unifying,zhang2024drone,ye2025gs} representation, we render the predicted edited SLat from the editing camera view and compare it with the edited reference image using both pixel-level and perceptual objectives~\cite{fu2023dreamsim}.

Let $\hat{z}^{e}_{\mathrm{slat}}$ denote the predicted edited SLat, $D_{\mathrm{GS}}(\cdot)$ the SLat-to-Gaussian decoder, $\mathcal{R}_{\mathrm{GS}}(\cdot,\pi)$ the Gaussian renderer under camera $\pi$, and $I^{e}_{\pi}$ the edited reference image. The rendered image is
\begin{equation}
\hat{I}_{\pi}
=
\mathcal{R}_{\mathrm{GS}}
\left(
D_{\mathrm{GS}}\left(\hat{z}^{e}_{\mathrm{slat}}\right),
\pi
\right).
\end{equation}
We define the render-space loss as
\begin{equation}
\mathcal{L}_{\mathrm{render}}
=
\mathbf{1}_{t<0.5}
\left(
\lambda_{\mathrm{mse}}
\left\|
\hat{I}_{\pi} - I^{e}_{\pi}
\right\|_2^2
+
\lambda_{\mathrm{ds}}
\mathcal{D}_{\mathrm{DreamSim}}
\left(
\hat{I}_{\pi},
I^{e}_{\pi}
\right)
\right).
\end{equation}
We apply this loss only when $t<0.5$, since the predicted SLat is sufficiently denoised at later flow steps for stable decoding and rendering. The MSE term encourages pixel-level alignment with the edited view, while the DreamSim term improves perceptual and semantic consistency.

The final objectives are stage-specific. Stage 1 optimizes only the sparse-structure flow loss and mask-aware velocity preservation:
\begin{equation}
\mathcal{L}_{\mathrm{stage1}}
=
\mathcal{L}_{\mathrm{flow}}^{\mathrm{SS}}
+
\lambda_{\mathrm{mask}}^{\mathrm{SS}}
\mathcal{L}_{\mathrm{mask}}^{\mathrm{SS}} .
\end{equation}
Stage 2 further adds the render-space consistency loss:
\begin{equation}
\mathcal{L}_{\mathrm{stage2}}
=
\mathcal{L}_{\mathrm{flow}}^{\mathrm{SLat}}
+
\lambda_{\mathrm{mask}}^{\mathrm{SLat}}
\mathcal{L}_{\mathrm{mask}}^{\mathrm{SLat}}
+
\lambda_{\mathrm{render}}
\mathcal{L}_{\mathrm{render}} .
\end{equation}
Thus, Stage 1 focuses on structure-level editing and preservation, while Stage 2 additionally enforces visual alignment with the editing condition.

\begin{table*}[t]
  \caption{Quantitative comparison the shape-edit split of Uni3DEdit-Bench.
  ``Train'' indicates whether the method requires training.
  ``3D Mask'' indicates whether the method requires a 3D edit-region mask at inference.}
  \label{tab:unifiedit_results}
  \centering
  \begin{tabular}{l|cc|ccc|cccc}
    \toprule
    \multirow{2}{*}{\textbf{Method}} &
    \multirow{2}{*}{\textbf{Train}} &
    \multicolumn{1}{c|}{\textbf{3D}} &
    \multicolumn{3}{c|}{\textbf{3D Metrics}} &
    \multicolumn{4}{c}{\textbf{2D Metrics}} \\
    &
    &
    \textbf{Mask} &
    CD$\downarrow$ &
    NC$\uparrow$ &
    F1$^{0.01}\uparrow$ &
    PSNR$\uparrow$ &
    SSIM$\uparrow$ &
    LPIPS$\downarrow$ &
    DINO-I$\uparrow$ \\
    \midrule

    Nano3D (ICLR'26)~\cite{ye2025nano3d}
    & \xmark & \xmark
    & 11.80 & 0.915 & 87.93
    & 28.69 & 0.958 & 0.043 & 0.952 \\

    VoxHammer (3DV'26)~\cite{li2025voxhammer}
    & \xmark & \cmark
    & 12.40 & 0.881 & 83.37
    & 24.66 & 0.920 & 0.071 & 0.917 \\

    3DEditFormer (ICML'26)~\cite{xia2025scalable3dediting}
    & \cmark & \xmark
    & 18.58 & 0.849 & 76.96
    & 24.26 & 0.913 & 0.074 & 0.930 \\

    \textbf{PartFlow}
    & \cmark & \xmark
    & \textbf{5.09} & \textbf{0.929} & \textbf{91.88}
    & \textbf{29.50} & \textbf{0.959} & \textbf{0.031} & \textbf{0.973} \\

    \bottomrule
  \end{tabular}
\end{table*}

\begin{table*}[t]
  \caption{Ablation study on the shape-edit split of Uni3DEdit-Bench.}
  \label{tab:ablation_study}
  \centering
  \begin{tabular}{l|ccc|cccc}
    \toprule
    \multirow{2}{*}{\textbf{Method}} &
    \multicolumn{3}{c|}{\textbf{3D Metrics}} &
    \multicolumn{4}{c}{\textbf{2D Metrics}} \\
    &
    CD$\downarrow$ &
    NC$\uparrow$ &
    F1$^{0.01}\uparrow$ &
    PSNR$\uparrow$ &
    SSIM$\uparrow$ &
    LPIPS$\downarrow$ &
    DINO-I$\uparrow$ \\
    \midrule

    \textbf{PartFlow}
    & \textbf{5.09} & \textbf{0.929} & \textbf{91.88}
    & \textbf{29.50} & \textbf{0.959} & \textbf{0.031} & \textbf{0.973} \\

    - Render-space Loss
    & \textbf{5.09} & 0.927 & 91.80
    & 27.96 & 0.945 & 0.042 & 0.964 \\

    - Mask Loss
    & 13.41 & 0.873 & 78.94
    & 26.48 & 0.931 & 0.055 & 0.957 \\
    
    \bottomrule
  \end{tabular}
\end{table*}

\begin{table}[t]
  \caption{Quantitative comparison on the appearance-edit split of Uni3DEdit-Bench.}
  \label{tab:appearance_results}
  \centering
  \begin{tabular}{l|cccc}
    \toprule
   \textbf{Method} &
    PSNR$\uparrow$ &
    SSIM$\uparrow$ &
    LPIPS$\downarrow$ &
    DINO-I$\uparrow$ \\
    \midrule

    TRELLIS~\cite{xiang2025structured}
    & 26.70 & 0.934 & 0.038 & 0.970 \\

    Nano3D~\cite{ye2025nano3d}
    & 27.09 & 0.938 & 0.036 & 0.971 \\

    VoxHammer~\cite{li2025voxhammer}
    & \textbf{28.40} & \textbf{0.950} & 0.033 & 0.966 \\

    \textbf{PartFlow}
    & 28.05 & 0.944 & \textbf{0.029} & \textbf{0.979} \\

    \bottomrule
  \end{tabular}
\end{table}

\section{Experiments}

\subsection{Implementation Details}

We evaluate our method on \textbf{Uni3DEdit-Bench}, a manually curated benchmark built from the Pxform test split. Uni3DEdit-Bench contains 1,497 high-quality editing cases across seven edit types. We divide these cases into two groups according to the nature of the edit. The \textit{shape-edit} split contains 1,265 samples, covering deletion, addition, local replacement, and local scaling. The remaining 232 samples form the \textit{appearance-edit} split, covering local color change and local material change.

For shape editing, we compare PartFlow with representative state-of-the-art methods, including the training-free method Nano3D~\cite{ye2025nano3d} and the training-based method 3DEditFormer~\cite{xia2025scalable3dediting}. For appearance editing, we additionally include TRELLIS~\cite{xiang2025structured} as a strong image-to-3D baseline, together with Nano3D and 3DEditFormer. All methods are evaluated using the same source assets, edit instructions, and rendering cameras when applicable. For fair comparison, we render all edited outputs from the same set of evaluation views and compute metrics on the aligned before/after renderings and reconstructed 3D outputs.

\subsection{Evaluation Metrics}

Following 3DEditVerse~\cite{xia2025scalable3dediting}, we evaluate shape edits using both 3D geometry metrics and multi-view rendering metrics.

\paragraph{Shape editing metrics.}
For 3D geometry evaluation, we uniformly sample 100,000 surface points from both the predicted mesh and the ground-truth mesh. We report Chamfer Distance (CD)~\cite{fan2017pointset}, Normal Consistency (NC)~\cite{gkioxari2019meshrcnn}, and F-score at threshold 0.01 (F1@0.01)~\cite{knapitsch2017tanks}. 

For render-based evaluation, we render each edited asset from 10 fixed camera views and compare the rendered images with the corresponding ground-truth edited renderings. We report PSNR, SSIM~\cite{wang2004ssim}, LPIPS~\cite{zhang2018lpips}, and DINO-I~\cite{oquab2023dinov2}. 

\paragraph{Appearance editing metrics.}
For local color change and local material change, the geometry should remain mostly unchanged while the appearance follows the edit instruction. We therefore evaluate appearance edits on the same 10 fixed views using PSNR, SSIM, LPIPS, and DINO-I between the rendered edited output and the ground-truth edited renderings.

\subsection{Quantitative Results}



Table~\ref{tab:unifiedit_results} reports quantitative results the shape-edit split of Uni3DEdit-Bench. PartFlow achieves the best performance across all 3D and 2D metrics. Compared with Nano3D, the strongest baseline in this table, PartFlow reduces CD from 11.80 to 5.09, improves NC from 0.915 to 0.929, and increases F1$^{0.01}$ from 87.93 to 91.88, demonstrating stronger geometric fidelity and source-structure preservation. PartFlow also improves the render-based metrics, increasing PSNR from 28.69 to 29.50, SSIM from 0.958 to 0.959, and DINO-I from 0.952 to 0.973, while reducing LPIPS from 0.043 to 0.031. Notably, PartFlow outperforms VoxHammer across all metrics without requiring a 3D edit-region mask at inference, showing better practicality for user-facing 3D editing.

Table~\ref{tab:appearance_results} evaluates the appearance-edit split. Compared with TRELLIS, PartFlow improves PSNR from 26.70 to 28.05, SSIM from 0.934 to 0.944, LPIPS from 0.038 to 0.029, and DINO-I from 0.970 to 0.979. Compared with Nano3D, PartFlow also achieves consistent gains across all four metrics. Although VoxHammer obtains slightly higher PSNR and SSIM, PartFlow achieves better LPIPS and DINO-I without relying on a 3D edit-region mask at inference. These results indicate that PartFlow provides stronger perceptual fidelity and semantic visual alignment while maintaining effective preservation for appearance-level editing.

\subsection{Qualitative Results}

Figure~\ref{fig:qualitative_shape} presents qualitative comparisons on shape editing. VoxHammer and Nano3D generally preserve the source asset well, but their condition-following ability is limited: they often leave the target region unchanged, perform incomplete edits, or modify the wrong structure. 3DEditFormer shows stronger responsiveness to shape-edit instructions, but it is less stable in source preservation and may introduce structural distortion or off-target changes. In contrast, PartFlow achieves both accurate local editing and strong preservation of the source identity, layout, and unedited regions. This benefits from the source-latent control and mask-aware preservation loss, which suppress unintended changes, while the render-space consistency loss further improves condition alignment. Trained on Pxform’s semantic-part transformations, PartFlow generalizes better across diverse shape-edit instructions and produces more reliable structure-preserving results.

Figure~\ref{fig:qualitative_ap} further compares appearance editing results. TRELLIS and Nano3D usually avoid severe geometric artifacts, but their edits are often not fully aligned with the target color or material instruction. VoxHammer benefits from mask-based preservation, but its 3D-mask-guided modification may introduce visible artifacts or unstable local appearance changes. PartFlow more accurately applies the requested color and material edits while keeping the geometry and unedited appearance regions stable. These results show that Pxform provides effective scalable supervision for appearance-level 3D editing, and that the render-space consistency loss strengthens visual fidelity and instruction alignment.

\subsection{Ablation Study}

Table~\ref{tab:ablation_study} analyzes the contribution of the render-space consistency loss and the mask-aware velocity preservation loss. Removing the render-space loss keeps the 3D geometry metrics nearly unchanged, but clearly degrades 2D visual metrics: PSNR drops from 29.50 to 27.96, SSIM decreases from 0.959 to 0.945, LPIPS increases from 0.031 to 0.042, and DINO-I decreases from 0.973 to 0.964. This confirms that the render-space loss mainly improves visual alignment with the editing condition.

Removing the mask-aware preservation loss causes a much larger degradation, especially in 3D geometry metrics. CD increases from 5.09 to 13.41, NC drops from 0.929 to 0.873, and F1$^{0.01}$ decreases from 91.88 to 78.94. The 2D metrics also decline consistently. These results indicate that mask-aware velocity preservation is critical for suppressing off-target changes and maintaining the source structure in unedited regions. Together, the two losses are complementary: the mask loss validates the role of semantic-part supervision in spatial preservation and geometry consistency, while the render-space loss strengthens visual controllability.

\section{Conclusion}
We showed that scalable feedforward 3D editing can be learned from semantic-part transformations. To this end, we introduced Pxform, a high-quality holistic 3D editing dataset constructed from part-semantic 3D assets with explicit region control, covering seven edit types across geometry, appearance, and global style changes. Built on Pxform, PartFlow learns source-preserving 3D edit transformations through source-latent control, training-only mask-aware velocity preservation, and render-space consistency supervision, while requiring no 3D edit mask at inference. We further introduced Uni3DEdit-Bench, a manually curated benchmark with 1,497 high-quality editing cases for evaluating shape and appearance editing. Experiments demonstrate that consistent semantic-part supervision, combined with source-aware control, is a key step toward practical, scalable, and general-purpose 3D editing.

\bibliographystyle{ACM-Reference-Format}
\bibliography{sample-base}

\appendix

\begin{figure*}[!t]
    \centering
    \includegraphics[width=1.0\textwidth]{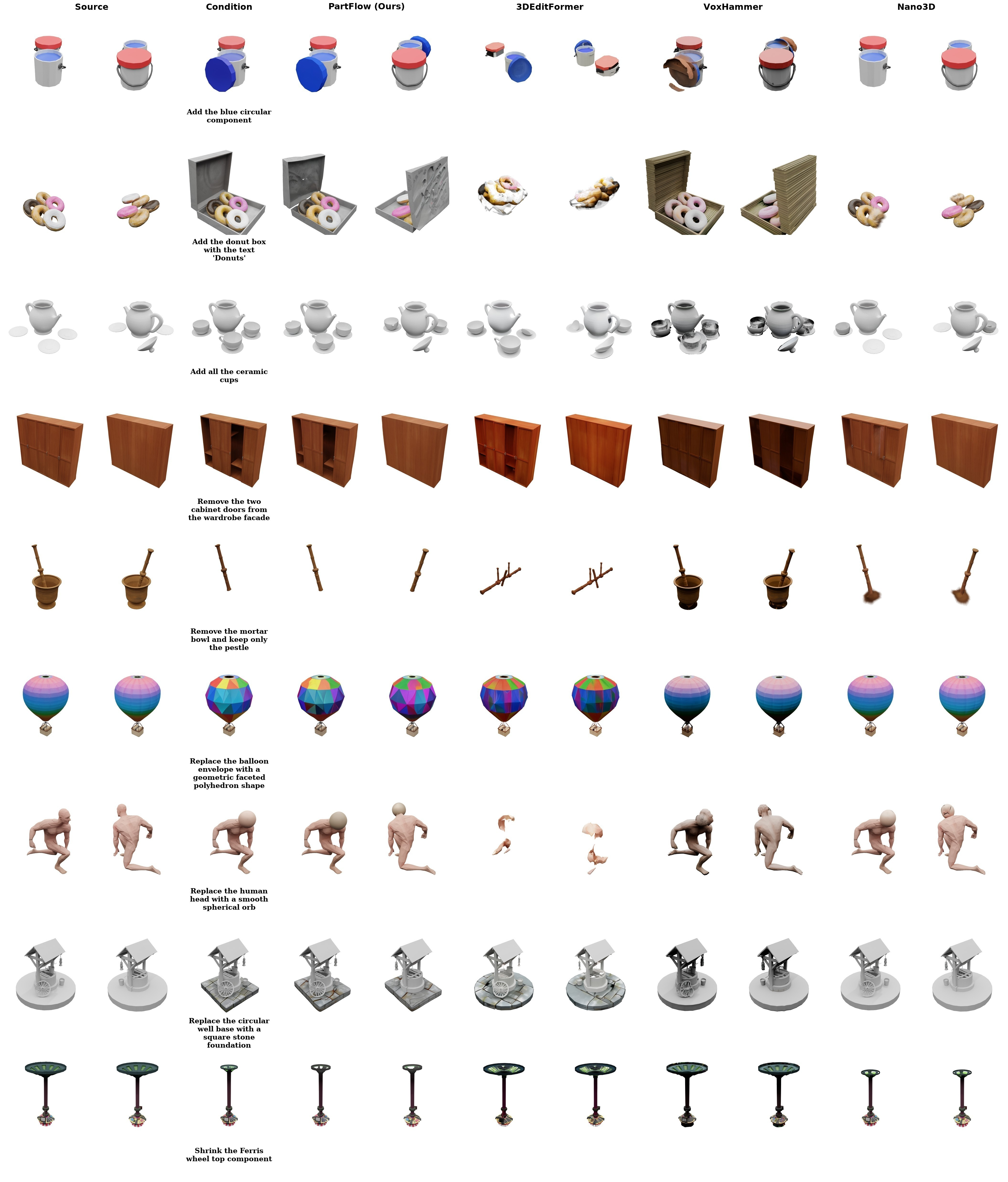}
    \caption{
        Qualitative results on Uni3DEdit-Bench for shape editing.
    }
    \label{fig:qualitative_shape}
\end{figure*}

\begin{figure*}[!t]
    \centering
    \includegraphics[width=1.0\textwidth]{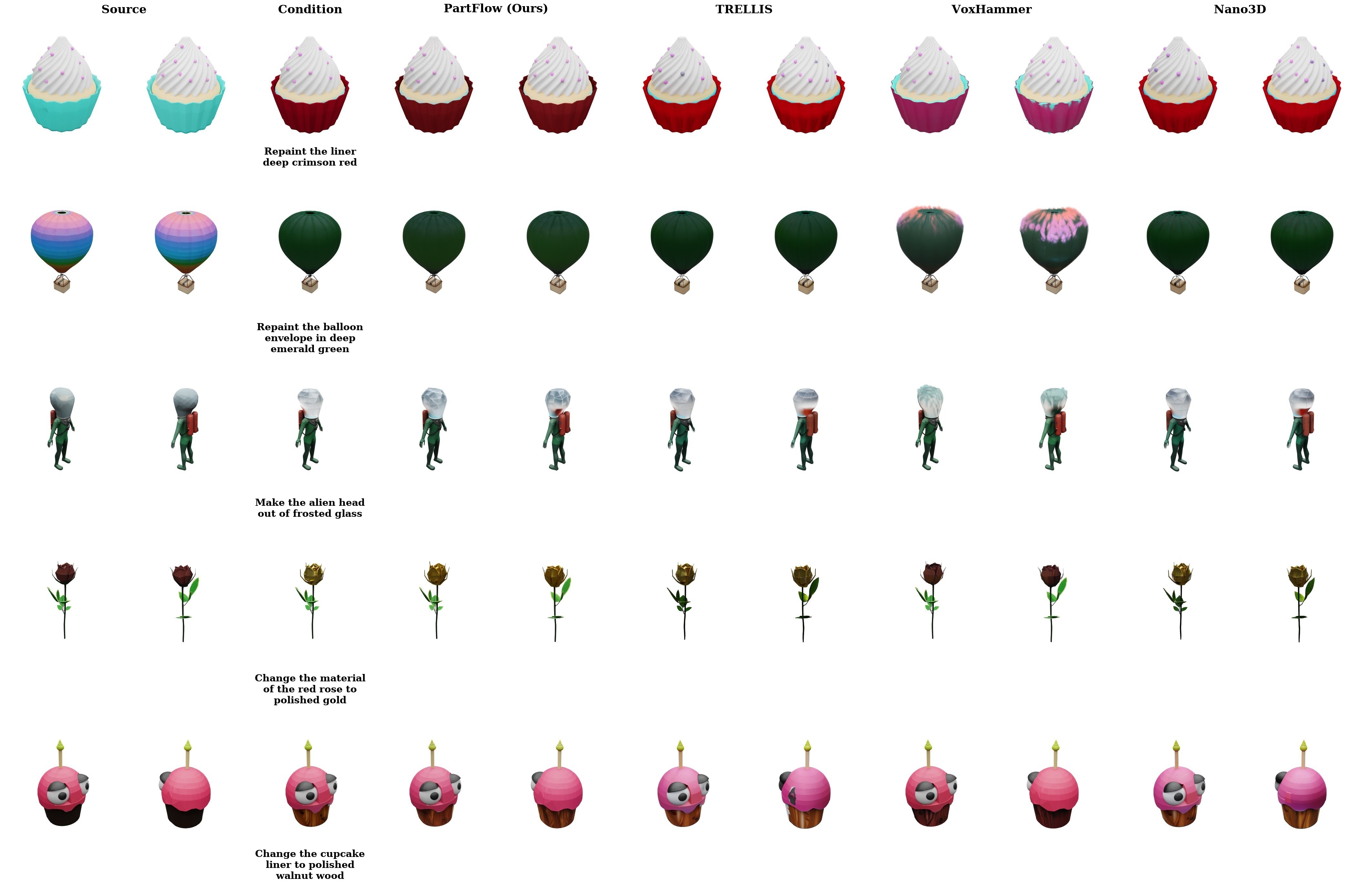}
    \caption{
        Qualitative results on Uni3DEdit-Bench for appearance editing.
    }
    \label{fig:qualitative_ap}
\end{figure*}

\begin{figure*}[!t]
    \centering
    \includegraphics[width=1.0\textwidth]{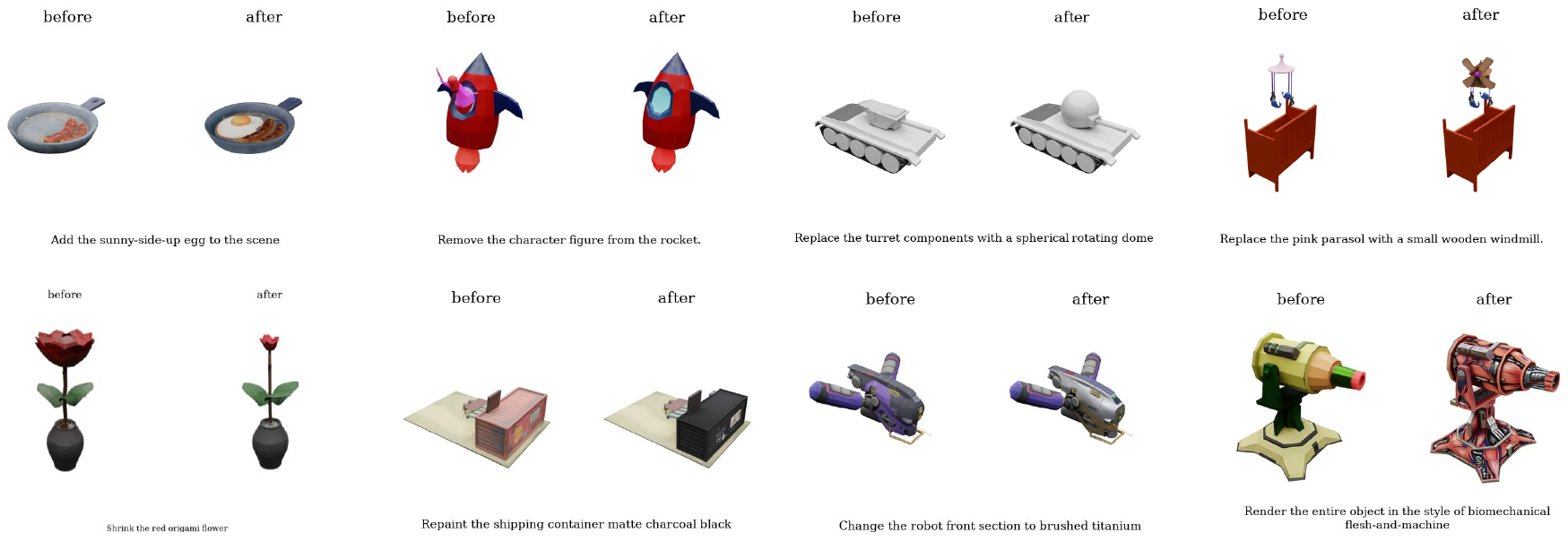}
    \caption{
        Some data from Pxform.
    }
    \label{fig:h3d_cases}
\end{figure*}

\clearpage
\onecolumn
\section{Implementation Details}
\subsection{Training Details}
PartFlow is trained in two stages following the TRELLIS pipeline. We train the Stage-1 sparse-structure model and the Stage-2 SLat model separately on 4 H200 GPUs with a batch size of 16. For each stage, we first optimize the standard flow-matching objective for 40K steps. We then continue training for another 5K steps with the proposed auxiliary losses enabled. Specifically, the mask-aware preservation loss is applied to strengthen source preservation in unedited regions, and the render-space consistency loss is introduced in Stage 2 to improve visual fidelity and condition alignment.

\subsection{Metrics}
CD measures the bidirectional nearest-neighbor distance between two point sets, NC evaluates the alignment of local surface normals, and F1@0.01 computes the harmonic mean of precision and recall under a strict distance threshold. Lower CD and higher NC/F1@0.01 indicate better geometric fidelity.

PSNR and SSIM measure pixel-level and structural similarity, LPIPS measures perceptual distance, and DINO-I computes feature-level similarity using DINOv2 image embeddings. Higher PSNR, SSIM, and DINO-I and lower LPIPS indicate better visual alignment.

\subsection{More Experiment}
To evaluate the generalization ability of our method, we further conduct experiments on Edit3D-Bench~\cite{ma2025steer3d}. For geometry evaluation, we sample 20,000 surface points from each generated and reference mesh to compute CD, and report F1 at a threshold of 0.01. As shown in Table~\ref{tab:edit3d_bench}, our method achieves the best average performance across add and remove editing tasks, with the lowest Avg CD and LPIPS, as well as the highest Avg F1. In particular, our method consistently improves CD and F1 on both add and remove tasks, indicating that it can produce more accurate geometric modifications while preserving the target structure. Although 3DEditFormer~\cite{xia2025scalable3dediting} obtains slightly lower LPIPS on the add task, our method achieves better overall appearance consistency on average and performs more robustly on the remove task. These results demonstrate that our method is not limited to our proposed benchmark, but can also generalize effectively to external 3D editing benchmarks.

\begin{table*}[t]
  \caption{Comparison on add and remove editing tasks in Edit3D-Bench.}
  \label{tab:edit3d_bench}
  \centering
  \begin{tabular}{l|ccc|ccc|ccc}
    \toprule
    \multirow{2}{*}{\textbf{Method}} &
    \multicolumn{3}{c|}{\textbf{Add}} &
    \multicolumn{3}{c|}{\textbf{Remove}} &
    \multicolumn{3}{c}{\textbf{Avg}} \\
    &
    LPIPS$\downarrow$ & CD$\downarrow$ & F1@1\%$\uparrow$ &
    LPIPS$\downarrow$ & CD$\downarrow$ & F1@1\%$\uparrow$ &
    LPIPS$\downarrow$ & CD$\downarrow$ & F1@1\%$\uparrow$ \\
    \midrule

    Steer3D~\cite{ma2025steer3d}
    & 0.2517 & 0.1351 & 0.2478
    & 0.2448 & 0.0862 & 0.3093
    & 0.2483 & 0.1107 & 0.2786 \\

    3DEditFormer~\cite{xia2025scalable3dediting}
    & \textbf{0.2407} & 0.0998 & 0.3300
    & 0.1343 & 0.0891 & 0.3100
    & 0.1875 & 0.0945 & 0.3200 \\

    PartFlow (Ours)
    & 0.2429 & \textbf{0.0925} & \textbf{0.3400}
    & \textbf{0.1223} & \textbf{0.0859} & \textbf{0.3380}
    & \textbf{0.1826} & \textbf{0.0892} & \textbf{0.3390} \\

    \bottomrule
  \end{tabular}
\end{table*}

\subsection{Limitations and Future Work}

Despite the effectiveness of Pxform and PartFlow, our current dataset does not yet support physically based rendering (PBR) materials. The appearance edits in Pxform focus on visually consistent color, material, and style transformations, but they do not explicitly model PBR attributes such as roughness, metallicity, normal maps, or physically grounded texture layers. This limits the dataset's applicability to production-level material editing and simulation-aware asset generation. In future work, we plan to extend the data construction pipeline toward PBR-aware 3D editing by collecting or synthesizing assets with structured material parameters and generating paired edits that preserve both visual realism and physically meaningful material properties.

Another limitation is that Pxform focuses on static 3D editing and does not include animation-level transformations. Some existing 3D datasets contain animated assets or motion-related annotations, and animation editing can also be handled by specialized motion generation and rigging-based methods. However, animation is not the central focus of this work. Our goal is to build a high-quality semantic-part transformation dataset for static geometry and appearance editing, where edit locality, source preservation, and before/after consistency are the main challenges. Nevertheless, extending Pxform to dynamic 3D assets is an important future direction. We plan to incorporate animation-related data, such as articulated motion, pose changes, and time-consistent appearance or geometry edits, to support more general 3D editing models in the future.
Looking further ahead, we hope to extend editing from static 3D objects to diverse 4D scenes \cite{liu2025light,chen2026one,shen2026gaussianart,huang2026arthoi,xu2025cruise}, where geometry, appearance, object states, and interactions evolve over time. Such 4D editing would require temporal consistency, physical plausibility, and controllable scene evolution, making semantic-part transformation data a potential foundation for learning world models that can understand, predict, and edit dynamic 3D environments.

\clearpage
\section{Details of Instructional Prompts}

Data engin pipeline uses one text-only LLM call for edit generation and two multimodal VLM calls for alignment and quality verification.

\subsection{Stage 1: Edit-Set Generation}

\begin{promptbox}{Stage 1 Prompt: Edit-Set Generation}
Role: You are a 3D-object edit-set generator. Given semantic object parts, generate a diverse and valid set of 3D edit instructions. Output exactly one JSON object only, with no prose and no Markdown fences.

Input: Text-only part list. Each item contains a part_id and a semantic caption. No image is provided in this stage. The user message also provides the required edit quota for each edit type and, when needed, a global-style roster sampled by object-specific variety seed.
Edit types: deletion / modification / scale / material / color / global
Decision rule: Assign exactly one edit_type to each edit. Do not conflate local geometry modification with material or color changes. Modification changes shape or identity; material changes surface substance or finish; color changes hue or shade only.

Task:
1. Generate exactly the requested number of edits according to the quota.
2. Select valid target parts from the supplied part list.
3. Group repeated instances into one edit when they form a semantic unit, e.g., both wheels or all chair legs.
4. Group semantically coupled components when appropriate, e.g., eyes, nose, and mouth as a head-level edit.
5. Avoid deleting the primary structural body of the object.
6. Use clear English imperative edit instructions.
7. Produce target descriptions and after-edit descriptions suitable for downstream editing and verification.

Rules:
R1. selected_part_ids must be a subset of the input part list.
R2. No two edits may share the same (edit_type, selected_part_ids) pair.
R3. Global edits must use selected_part_ids = [].
R4. Deletion cannot target the primary structural body.
R5. Modification may target the primary structural body if semantically valid.
R6. Internal identifiers such as part_3 must not appear in user-facing descriptions.
R7. All user-facing strings must be written in English.

Per-type edit_params schema:
deletion     -> {}
modification -> {"new_part_desc": "<geometry + identity>"}
scale        -> {"factor": <0.3-0.85>}
material     -> {"target_material": "<surface substance>"}
color        -> {"target_color": "<hue + shade phrase>"}
global       -> {"target_style": "<artistic style>"}

User template:
# PART LIST (id - description)
{part_menu}
Generate EXACTLY {n_total} edits:
{n_deletion} deletion, {n_modification} modification, {n_scale} scale,
{n_material} material, {n_color} color, {n_global} global{global_note}
If global edits are requested, {global_note} provides a sampled style roster.
Select global styles from different style categories when multiple global edits are required.
Output schema:
{
  "object": {
    "full_desc": "<English object description>",
    "parts": [{"part_id": int, "name": "<semantic label>"}]
  },
  "edits": [{
    "edit_type": "deletion|modification|scale|material|color|global",
    "selected_part_ids": [int],
    "prompt": "<imperative edit instruction>",
    "target_part_desc": "<description of target region>",
    "edit_params": {...},
    "after_desc": "<post-edit description; null for deletion>"
  }]
}
Validation: The response must be parseable as one JSON object. Invalid JSON or schema violations trigger a bounded retry with reduced quota.
\end{promptbox}

\subsection{Stage 2: Edit--Region Alignment Verification}

\begin{promptbox}{Stage 2 Prompt: Alignment Gate}
Role: You are a 3D-edit alignment judge. Determine whether the highlighted region matches the target part described by the edit instruction.

Input: For local edits, the image is a 5 x 2 grid: RGB views on top and highlight maps below. Selected parts are red, other parts are grey, and background is white. For global edits, the input is a 5 x 1 RGB strip and the whole object is treated as the target.

Text fields:
edit_type: <edit_type>
instruction: "<prompt>"
selected_part_ids: <selected_part_ids> or (none)
target_part_desc: "<target_part_desc>"

Task: Locate the red region, identify the corresponding physical component, compare it with the instruction and target description, decide semantic alignment, and select the best editing view.

Rules:
R1. aligned = true only if the highlighted region matches the named target and the instruction is clear.
R2. aligned = false for wrong selection, over-selection, under-selection, missing target, or incoherent instruction.
R3. best_view is the column where the target is most visible and suitable for editing.

Output fields:
aligned; reason; best_view.
\end{promptbox}

\subsection{Stage 3: Before/After Quality Judgement}

\begin{promptbox}{Stage 3 Prompt: Quality Gate}
Role: You are a 3D-edit quality judge. Given before/after multi-view renderings and edit metadata, judge whether the edit was correctly executed while preserving unrelated regions.

Visual input: A 2 x 5 collage of one object. The top row shows BEFORE views from cameras 0-4, and the bottom row shows AFTER views from the same cameras.

Text input:
edit_type: <edit_type>
Object: <object_desc>
Edit instruction: "<edit_prompt>"
Target part: <part_label>
Target description: "<target_part_desc>"
Expected edit attribute: <type-specific target>

Task: Localize the target region, compare same-camera BEFORE/AFTER pairs, verify whether the required edit is visible and consistent, check preservation of unrelated parts, penalize artifacts or wrong-region edits, and rewrite the prompt if the edit passes.

Expected change:
deletion -> target removed
modification/scale -> target geometry changed
material/color -> target appearance changed
global -> whole-object style changed

Hard-fail rules:
R1. Fail if BEFORE and AFTER are visually identical.
R2. Fail if the edit is missing or applied to the wrong region.
R3. Fail if geometry edits become appearance-only edits.
R4. Fail if color/material edits alter geometry.
R5. Fail if global edits destroy recognizable structure.

Output fields:
edit_executed; correct_region; preserve_other; visual_quality;
artifact_free; reason; prompt_quality; improved_prompt; improved_after_desc.
\end{promptbox}

\section{Additional Qualitative Results}

\begin{figure*}[!t]
    
    \centering
    \includegraphics[width=0.98\textwidth]{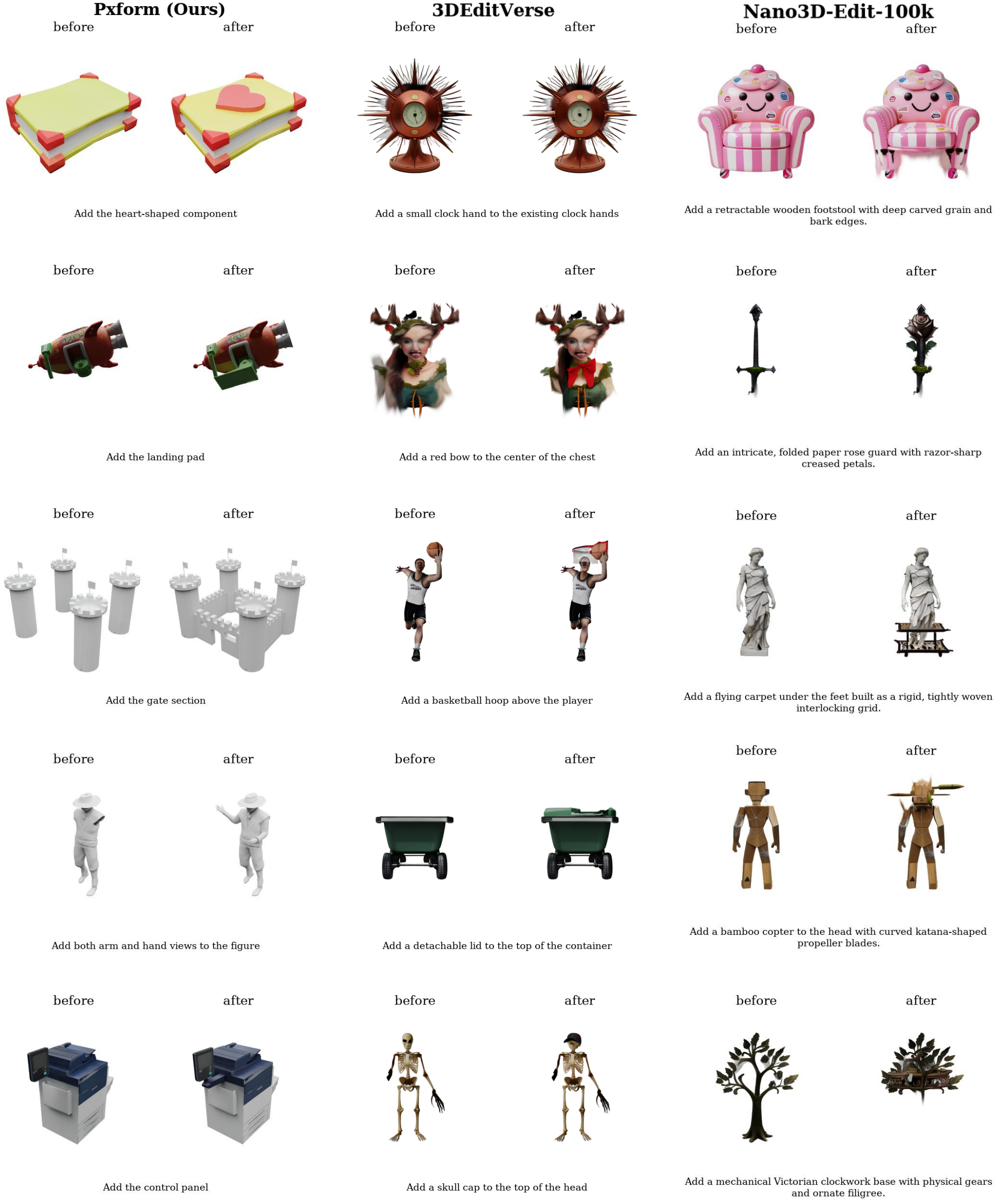}
    \caption{
       Qualitative comparison of Pxform, 3DEditVerse and Nano3D-Edit100k.
    }
    \label{fig:spc1}
\end{figure*}

\begin{figure*}[!t]
    \centering
    \includegraphics[width=0.98\textwidth]{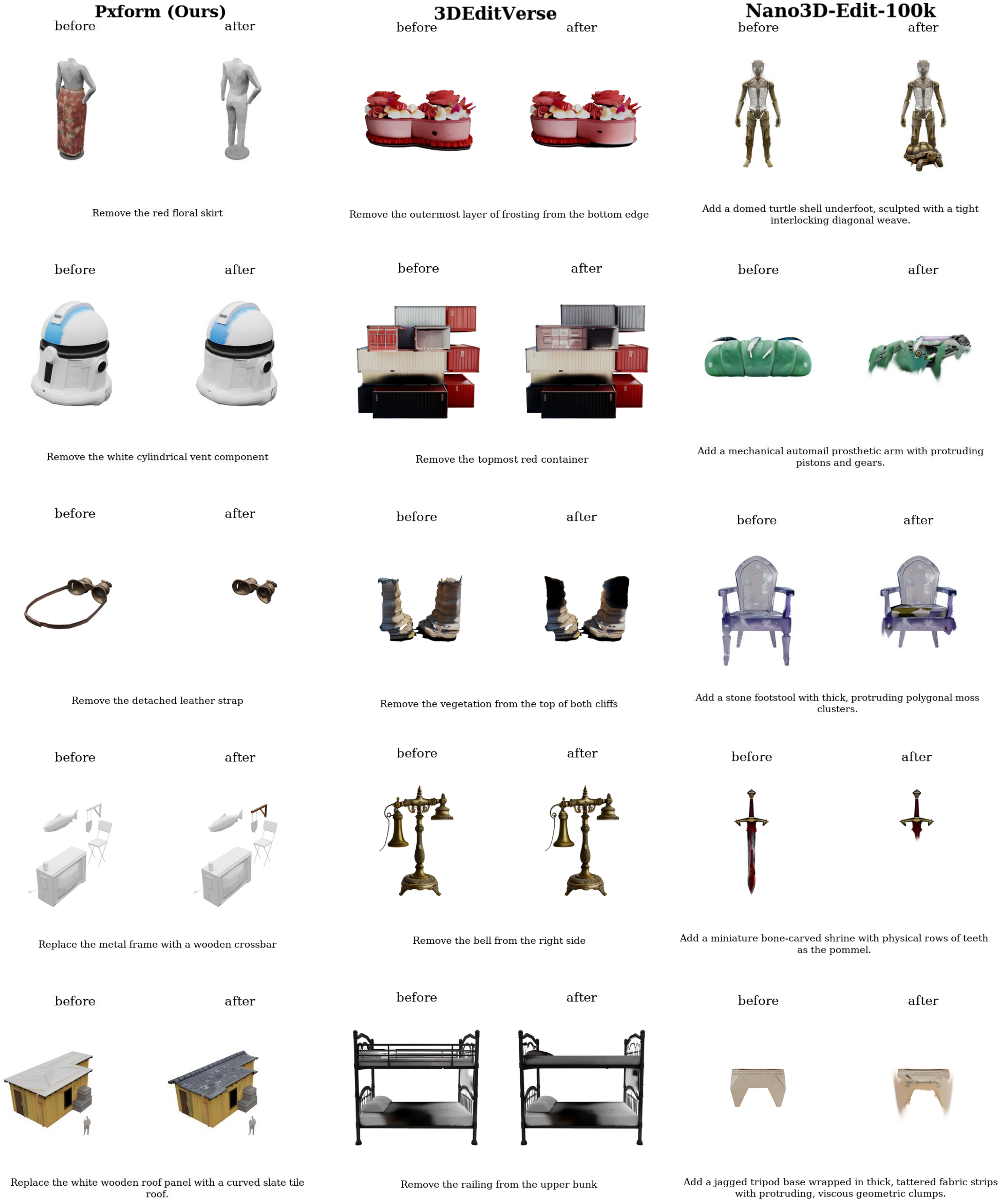}
    \caption{
        Qualitative comparison of Pxform, 3DEditVerse and Nano3D-Edit100k.
    }
    \label{fig:spc2}
\end{figure*}

\begin{figure*}[!t]
    \centering
    \includegraphics[width=0.98\textwidth]{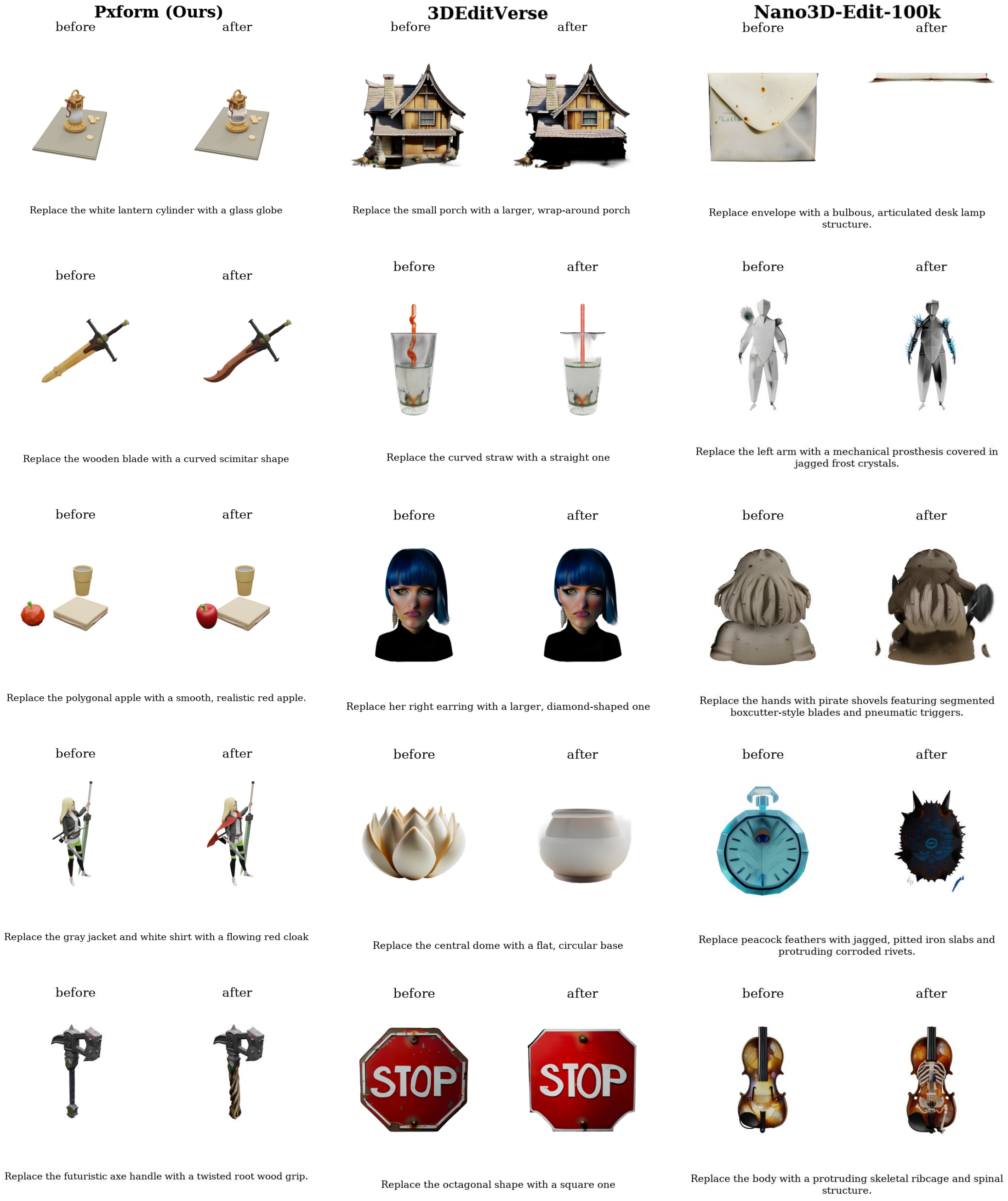}
    \caption{
        Qualitative comparison of Pxform, 3DEditVerse and Nano3D-Edit100k.
    }
    \label{fig:spc3}
\end{figure*}

\begin{figure*}[!t]
    \centering
    \includegraphics[width=\textwidth]{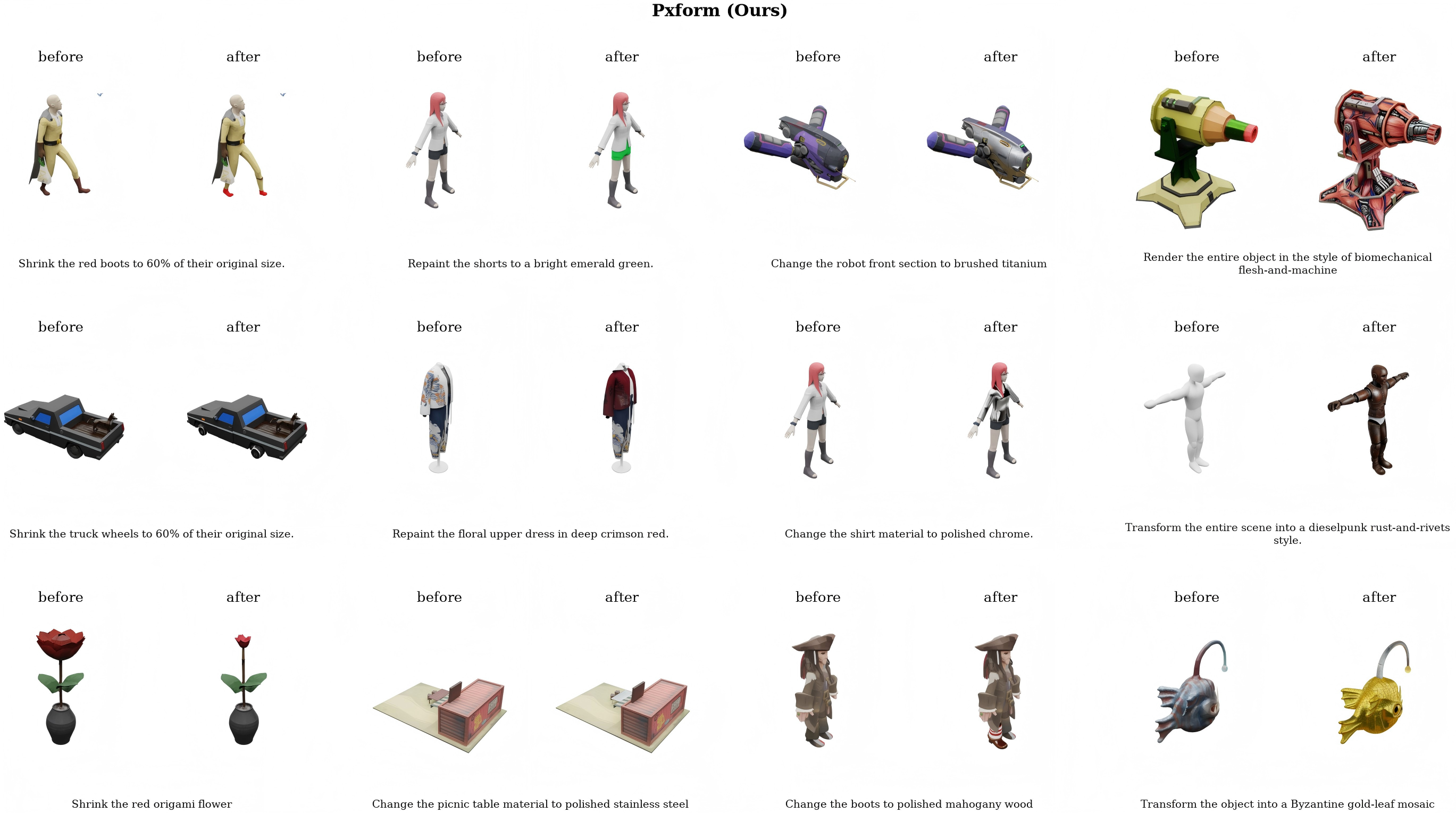}
    \caption{
        More Samples in Pxform Dataset.
    }
    \label{fig:cmgs}
\end{figure*}

\begin{figure*}[!htbp]
    \centering
    \includegraphics[width=0.97\textwidth]{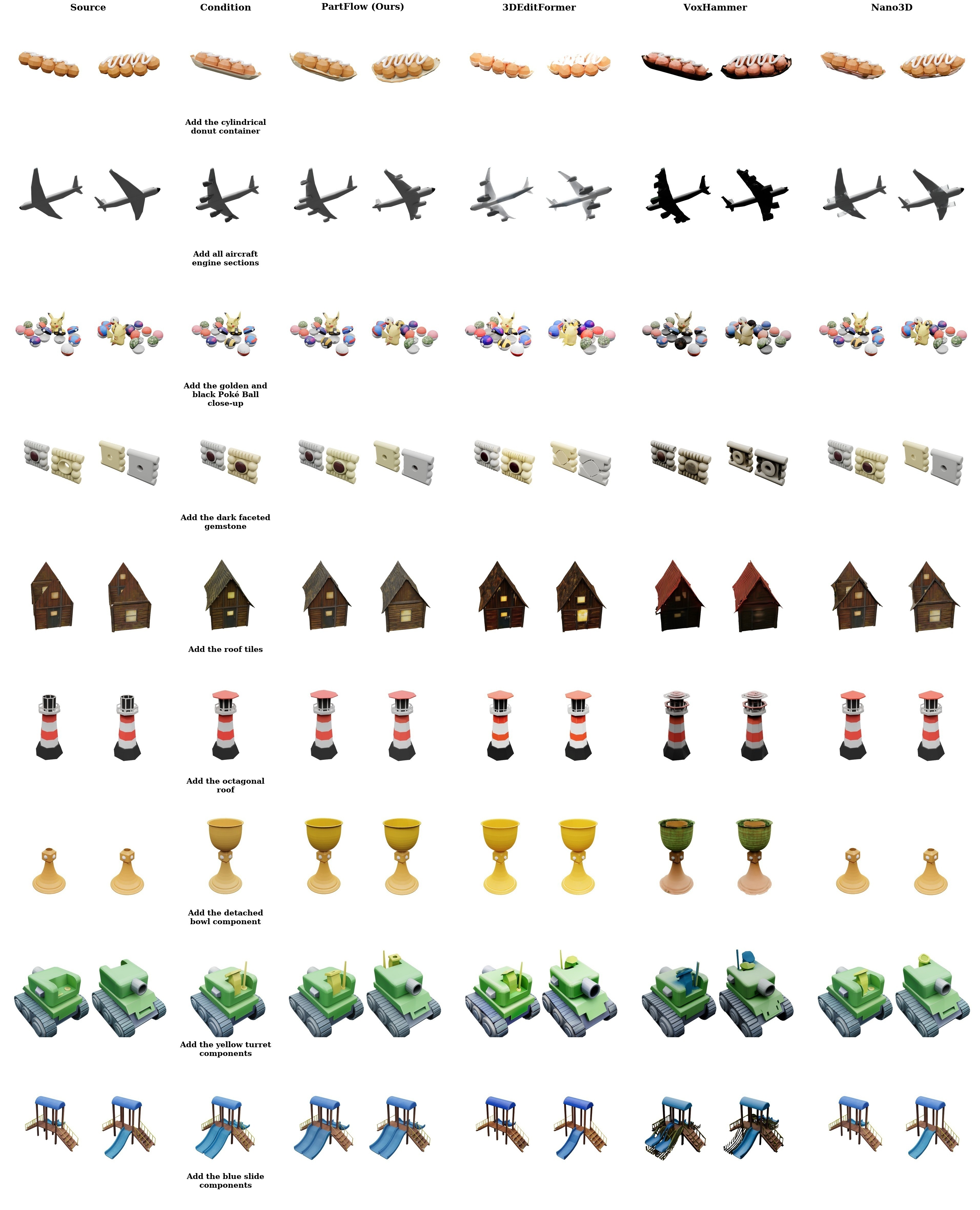}
    \caption{
        Qualitative results on Uni3DEdit-Bench.
    }
    \label{fig:2-1}
\end{figure*}

\begin{figure*}[!htbp]
    \centering
    \includegraphics[width=0.97\textwidth]{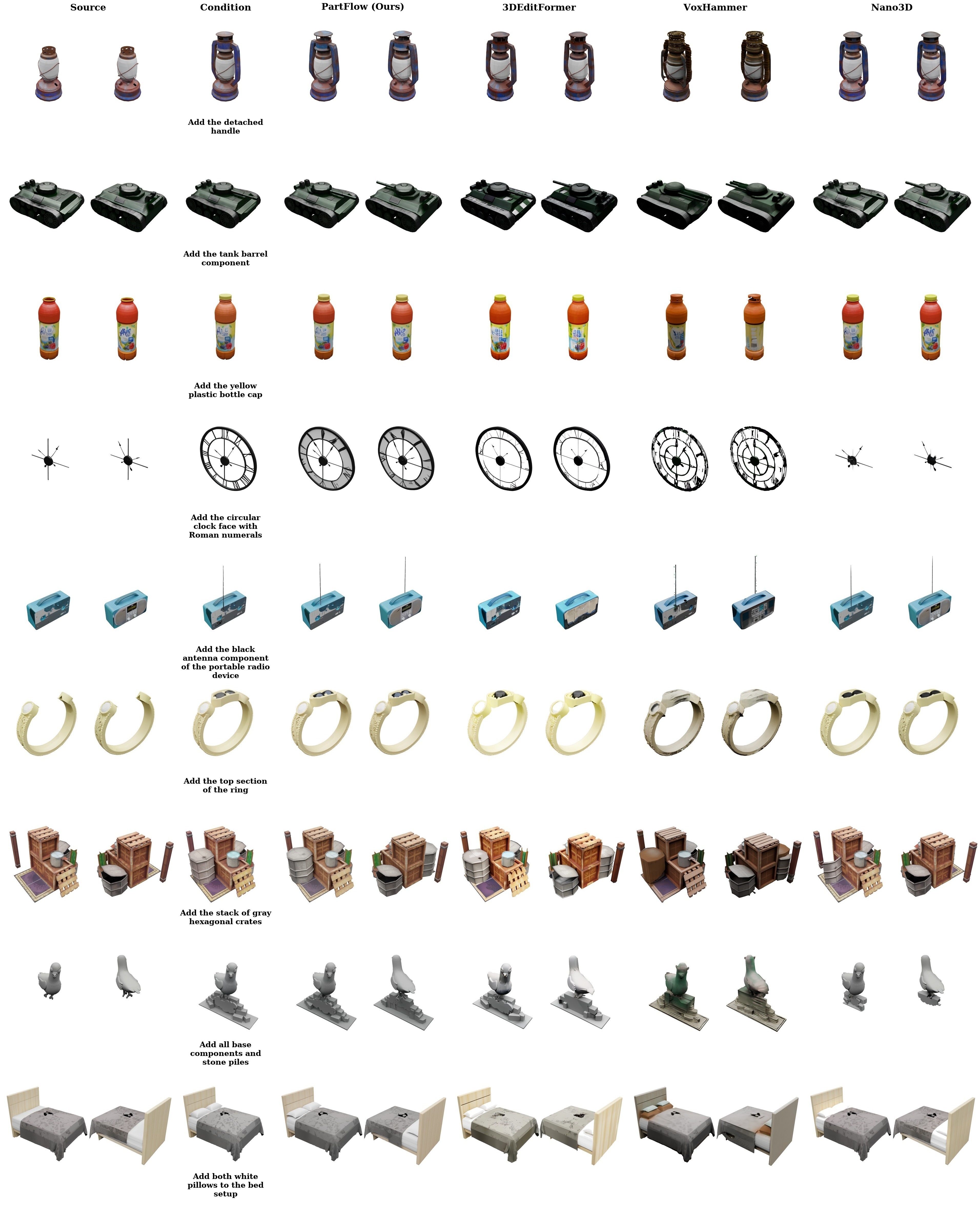}
    \caption{
        Qualitative results on Uni3DEdit-Bench.
    }
    \label{fig:2-2}
\end{figure*}

\begin{figure*}[!htbp]
    \centering
    \includegraphics[width=\textwidth]{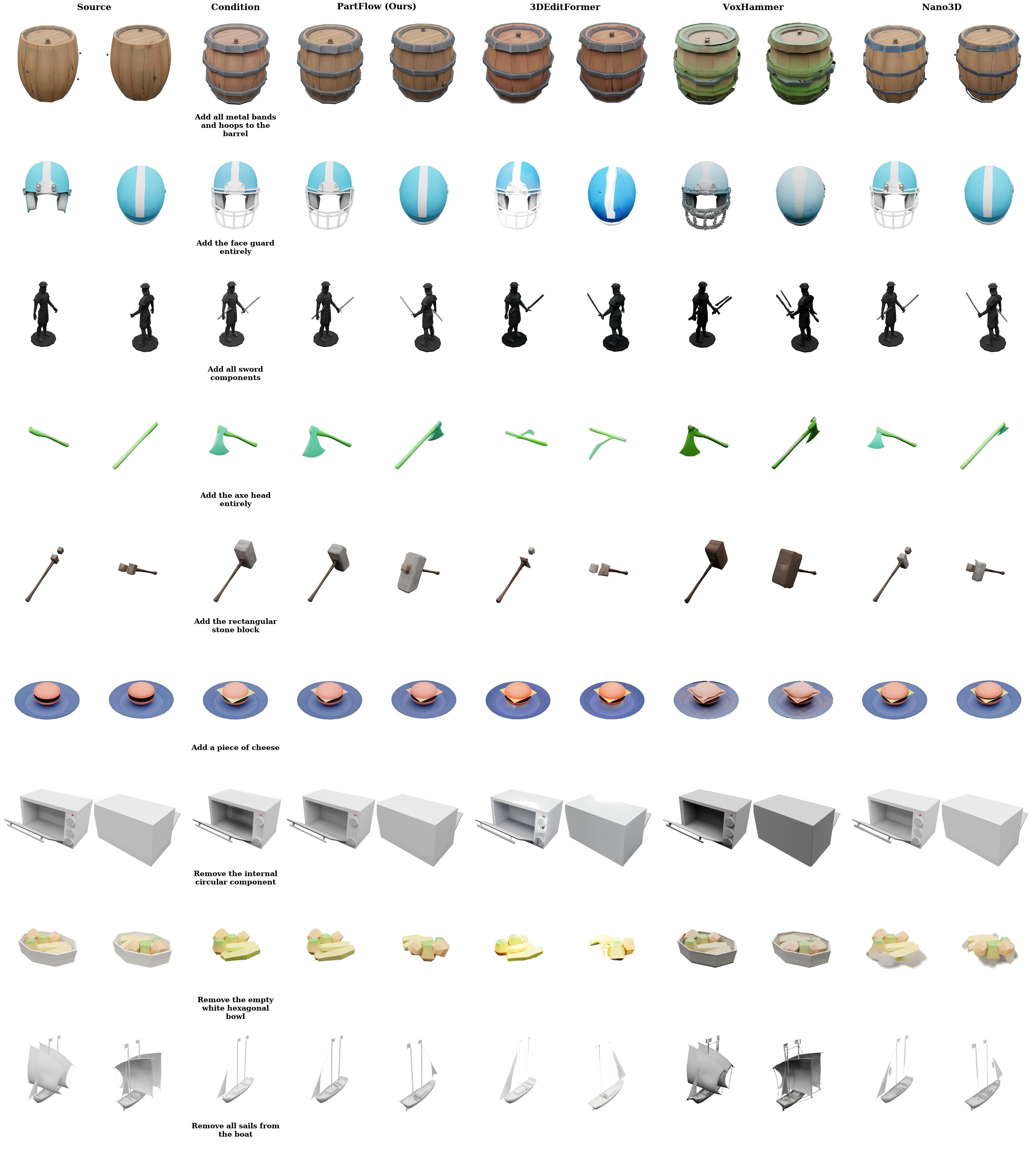}
    \caption{
        Qualitative results on Uni3DEdit-Bench.
    }
    \label{fig:2-3}
\end{figure*}

\begin{figure*}[!htbp]
    \centering
    \includegraphics[width=\textwidth]{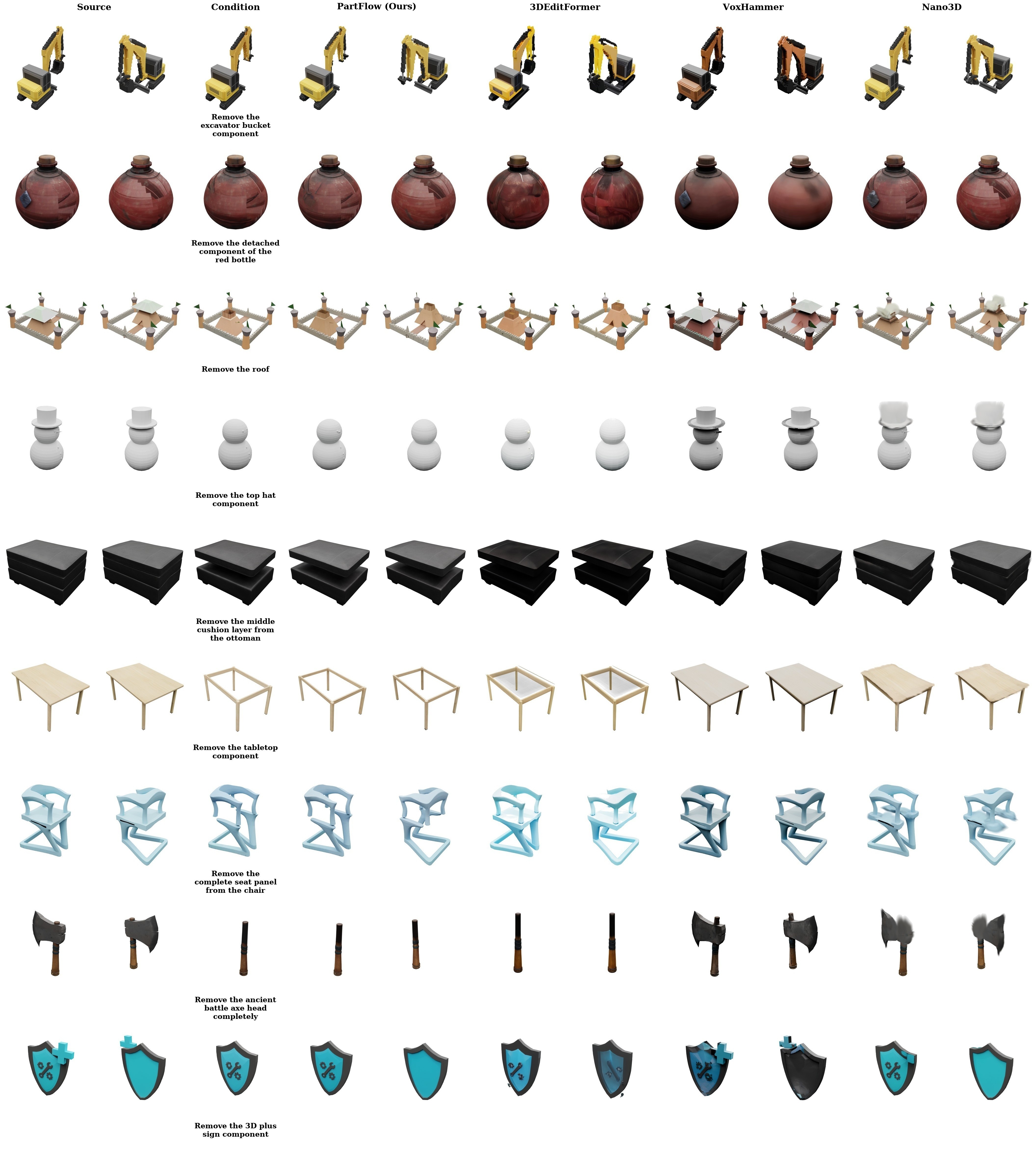}
    \caption{
        Qualitative results on Uni3DEdit-Bench.
    }
    \label{fig:2-4}
\end{figure*}

\begin{figure*}[!htbp]
    \centering
    \includegraphics[width=\textwidth]{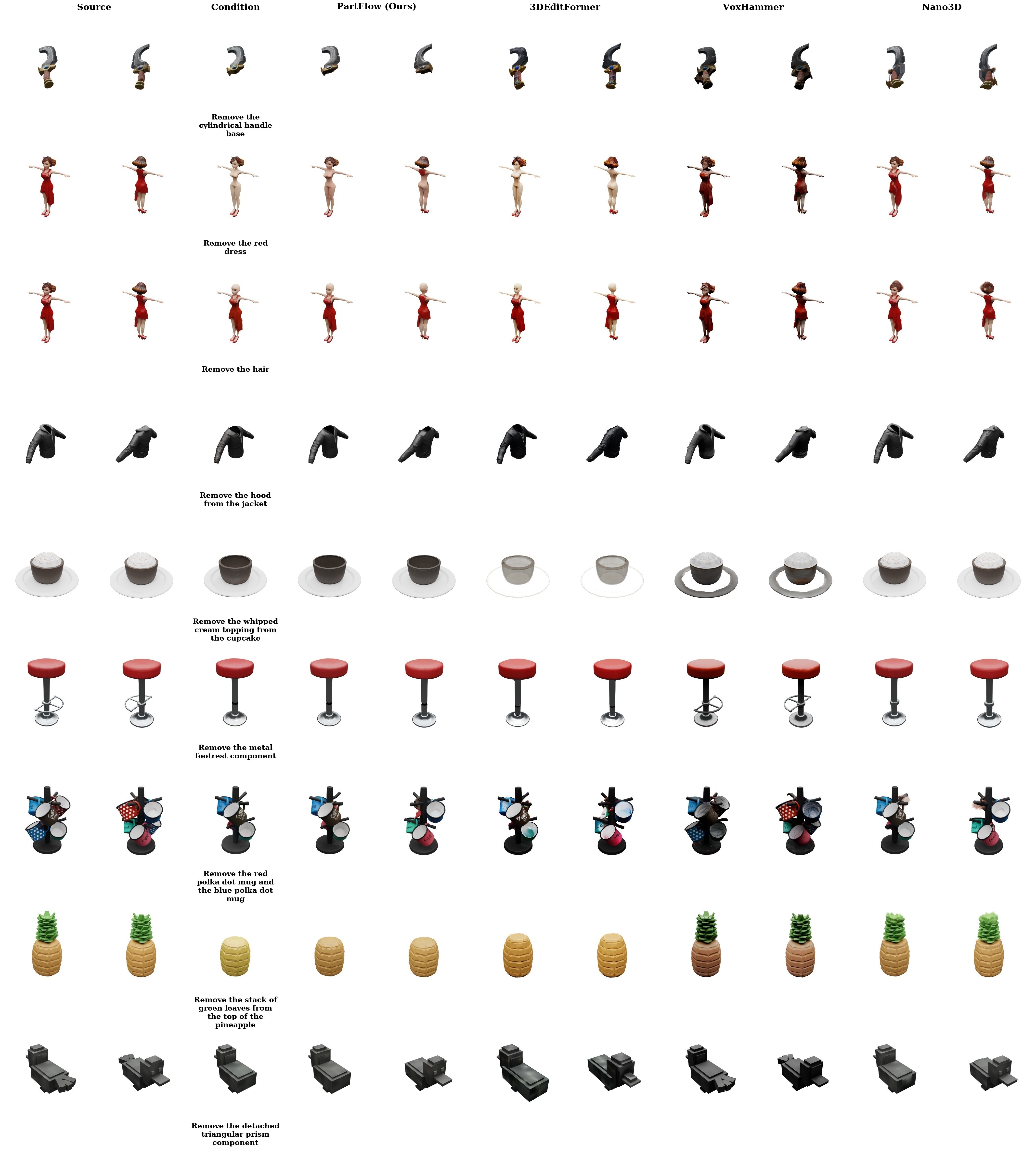}
    \caption{
        Qualitative results on Uni3DEdit-Bench.
    }
    \label{fig:2-5}
\end{figure*}

\begin{figure*}[!htbp]
    \centering
    \includegraphics[width=\textwidth]{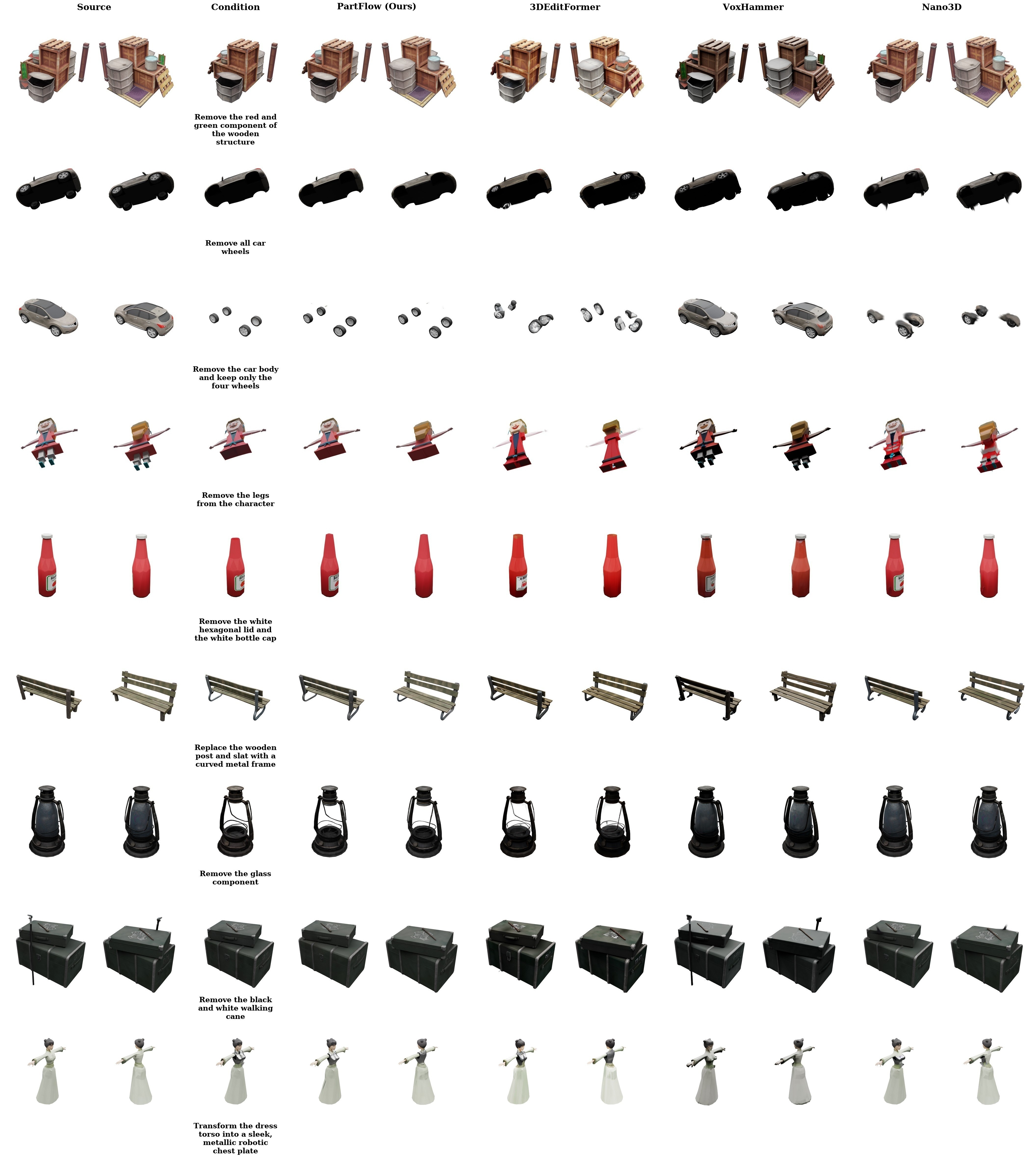}
    \caption{
        Qualitative results on Uni3DEdit-Bench.
    }
    \label{fig:2-6}
\end{figure*}

\begin{figure*}[!htbp]
    \centering
    \includegraphics[width=\textwidth]{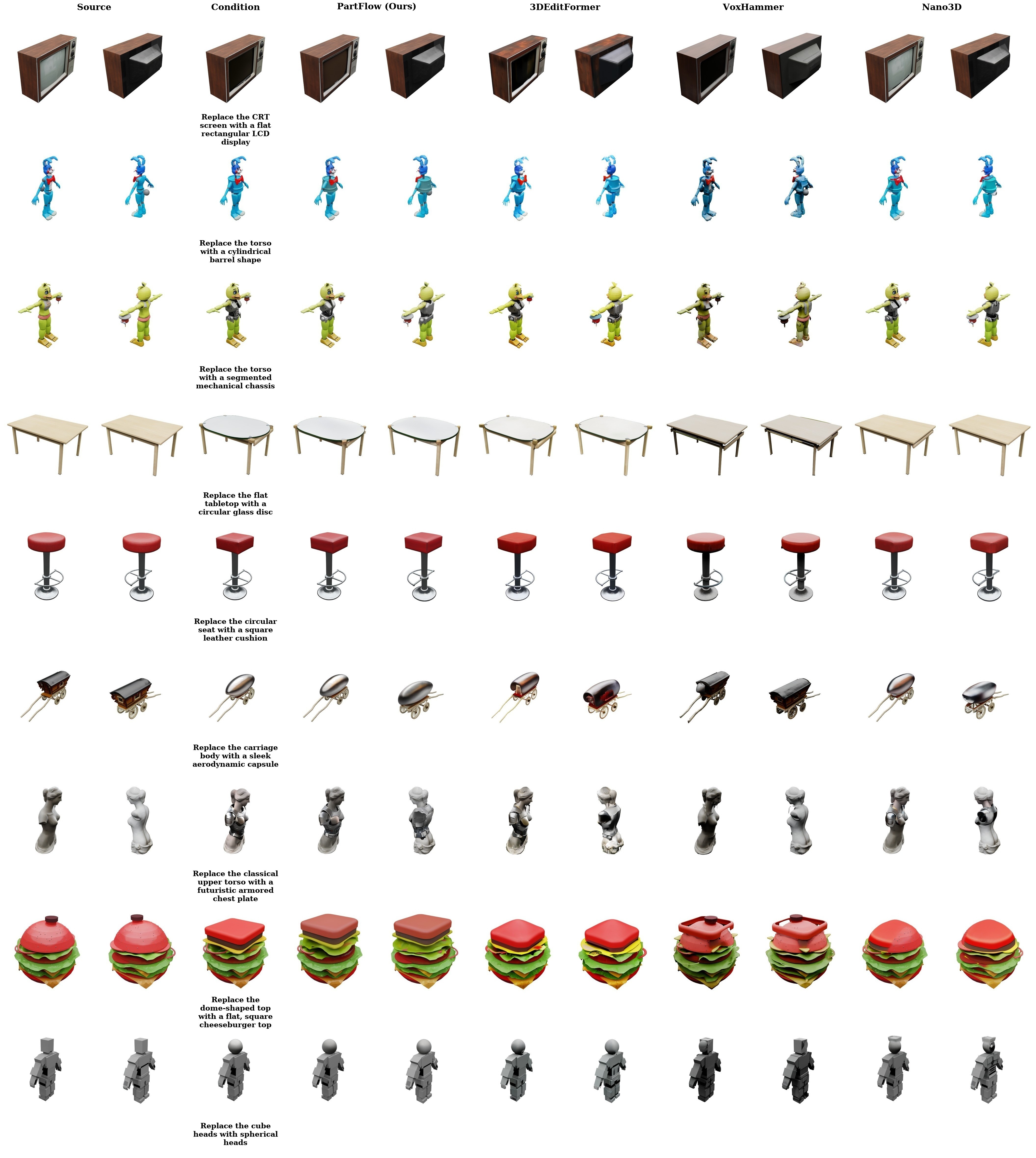}
    \caption{
        Qualitative results on Uni3DEdit-Bench.
    }
    \label{fig:2-7}
\end{figure*}

\begin{figure*}[!htbp]
    \centering
    \includegraphics[width=\textwidth]{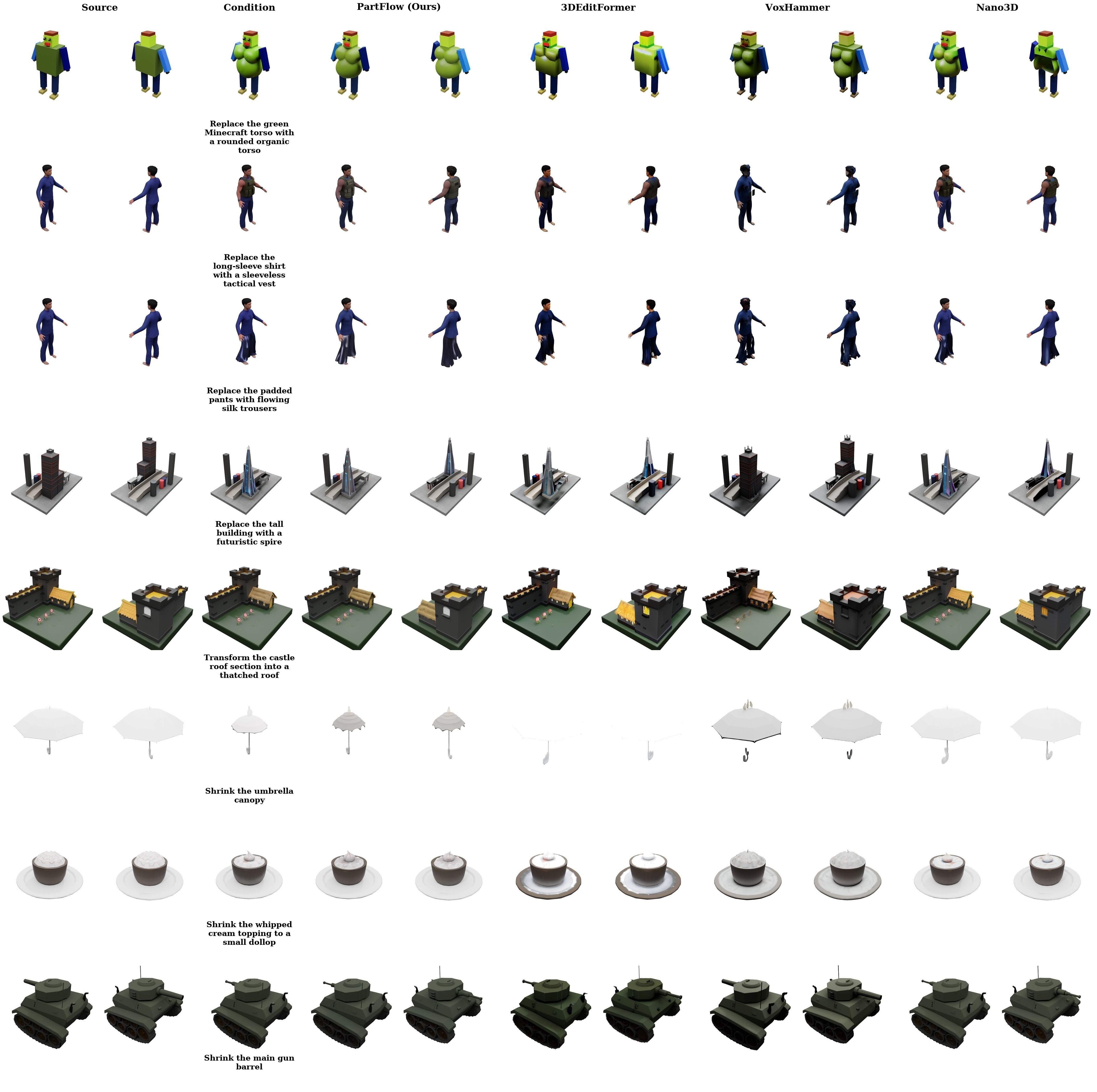}
    \caption{
        Qualitative results on Uni3DEdit-Bench.
    }
    \label{fig:2-8}
\end{figure*}

\begin{figure*}[!htbp]
    \centering
    \includegraphics[width=\textwidth]{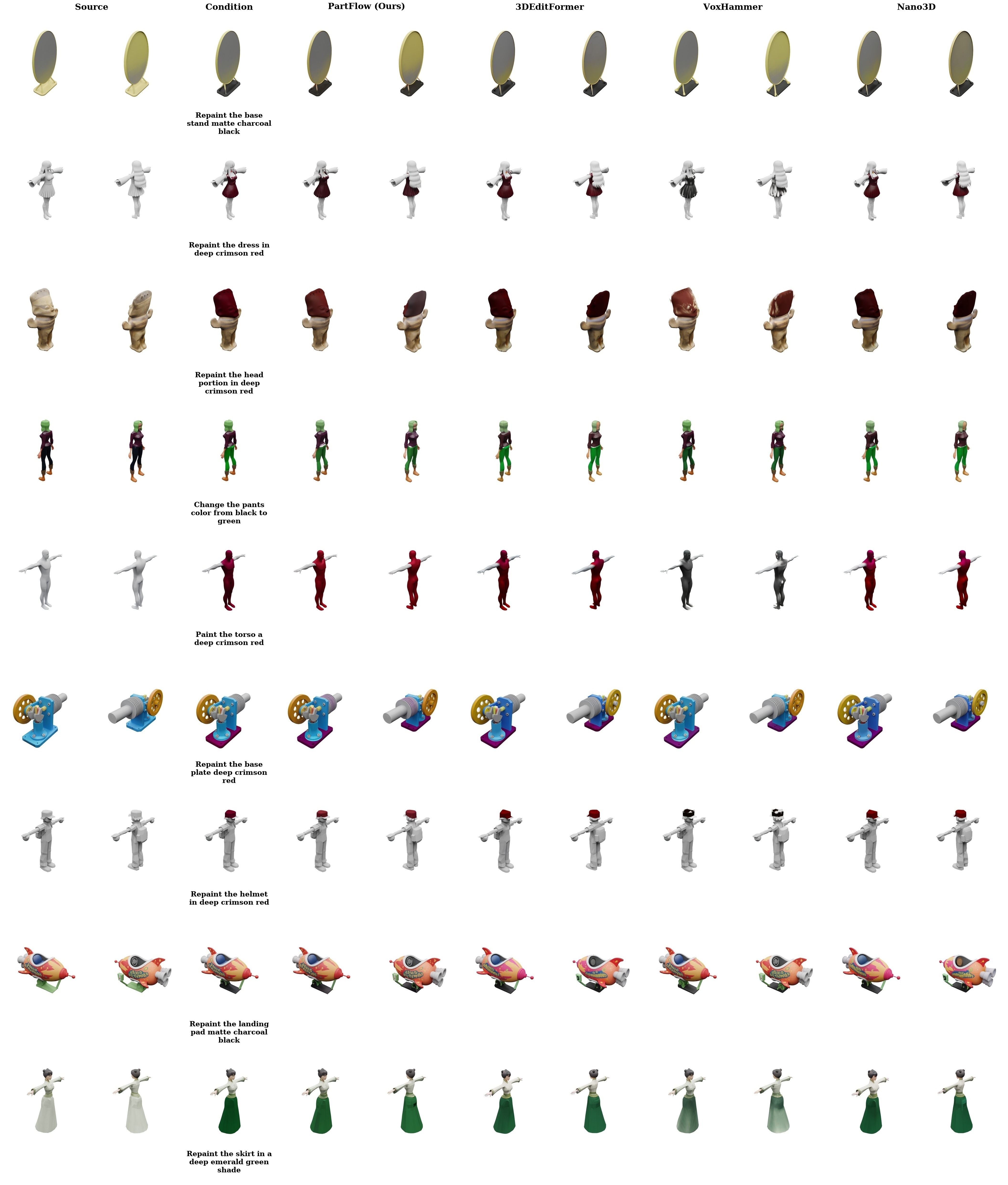}
    \caption{
        Qualitative results on Uni3DEdit-Bench.
    }
    \label{fig:2-9}
\end{figure*}

\begin{figure*}[!htbp]
    \centering
    \includegraphics[width=\textwidth]{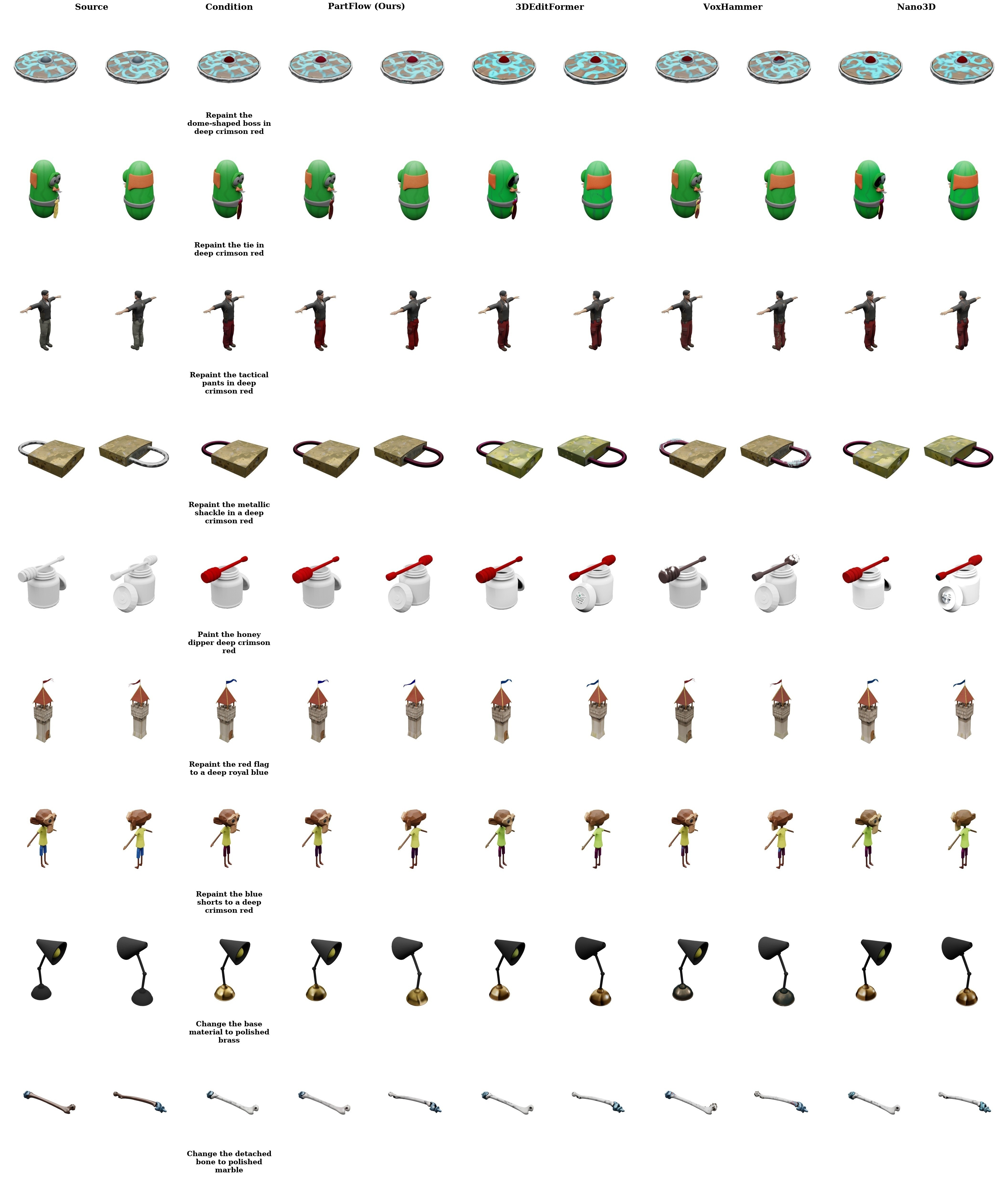}
    \caption{
        Qualitative results on Uni3DEdit-Bench.
    }
    \label{fig:2-10}
\end{figure*}
\begin{figure*}[!htbp]
    \centering
    \includegraphics[width=\textwidth]{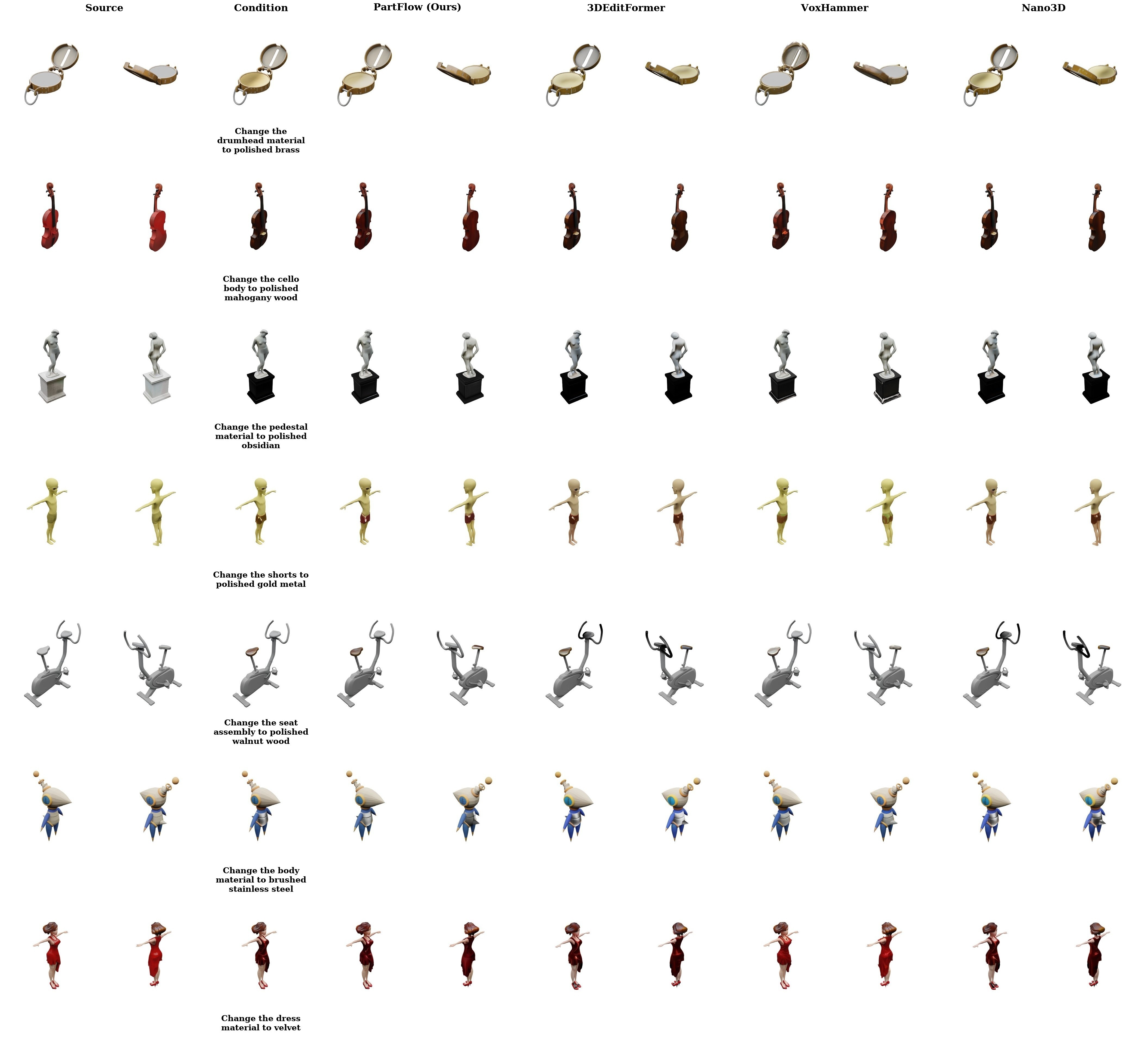}
    \caption{
        Qualitative results on Uni3DEdit-Bench.
    }
    \label{fig:2-11}
\end{figure*}

\end{document}